\documentclass[letterpaper, 10 pt, journal, twoside]{IEEEtran}
\usepackage{amsmath,amsfonts}
\usepackage{amssymb}
\usepackage{algorithmic}
\usepackage{algorithm}
\usepackage{array}
\usepackage[caption=false,font=footnotesize]{subfig}
\usepackage{stfloats}
\usepackage{textcomp}
\usepackage{stfloats}
\usepackage{url}
\usepackage{xcolor}
\usepackage{multirow}
\usepackage{verbatim}
\usepackage{graphicx}
\usepackage{makecell}
\usepackage{tabularx}
\usepackage{booktabs}
\usepackage{multirow}
\usepackage{float}

\usepackage{amssymb} 
\newcommand{\CheckmarkBold}{\ensuremath{\boldsymbol{\checkmark}}}
\newcommand{\CircleOpen}{\ensuremath{\circ}}
\newcommand{\NA}{---}
\newcolumntype{Y}{>{\centering\arraybackslash}X}

\usepackage{hyperref}
\usepackage[numbers,sort]{natbib}

\newcommand*{\vertbar}{\rule[-1ex]{0.5pt}{2.5ex}}

\newcommand{\di}[1]{\mathrm{d}#1}

\DeclareMathOperator*{\argmin}{\arg\!\min}

\begin{document}

\bstctlcite{IEEEexample:BSTControl}

\title{Real-Time Shape Control of Multi-Segment Soft Robotic Arms Using Koopman Operators with Global and Local Observables}


\author{Jiahe Wang, Eron Ristich, Sultan Haidar Ali, Eric Weissman, Lei Zhang, Wanxin Jin, Yi Ren, Jiefeng Sun*}





\maketitle

\begin{abstract}

Multi-segment soft robotic arms can continuously reconfigure their body shapes for safe interaction, but tip control alone is insufficient for constrained-space tasks. Therefore, shape control is a more important task for multi-segment soft arms than tip control, but remains challenging due to the high dimensionality and nonlinear dynamics of continuum deformation. In existing work, shape control accuracy is defined by the error in the global frame (global shape error). For multi-segment soft arms, using only global shape error as the control objective is insufficient, as segment coupling, gravity-induced loading, and inertial effects become more significant. This difficulty increases with the number of segments. 
In this paper, we present a Koopman-based model predictive control framework that combines global and local observables, enabling real-time shape control on multi-segment soft robotic arms. The framework is evaluated through numerical and physical experiments. Numerical experiments demonstrate the scalability of the proposed controller by achieving shape control on robots with up to 10 independently actuated segments. The physical experiments demonstrate that the controller is capable of 
(1) real-time shape control of 3- and 5-segment robotic arms with tip speeds up to 0.6 m/s, 
(2) robust tracking without retraining, including distal payloads up to 400~g and recovery from a 7~N lateral disturbance, and
(3) the potential for future inspection applications through a confined-space demonstration. These results demonstrate that the proposed framework enables dynamic, scalable, and accurate real-time shape control on multi-segment soft robotic arms.

\end{abstract}


\section{Introduction}


%









\IEEEPARstart{S}{oft} robotic arms, characterized by continuous morphology and intrinsic compliance, have emerged as a promising class of manipulators capable of safe and adaptive interaction with complex environments \cite{rus_design_2015, kulkarni_soft_2025}. 
They are particularly well-suited for applications such as minimally invasive surgery \cite{iqbal_continuum_2025}, cluttered environment operation \cite{coad_vine_2020}, and human–robot interaction \cite{zuo_umarm_2025}, where compliance and the ability to conform to complex geometries are essential.

Tip (end-effector) position and orientation control has been the primary focus of soft robotic arms research \cite{yu_data-efficient_2025, li_position_2025,lai_constrained_2022}. Various control strategies have been developed to improve the accuracy and robustness of tip tracking under payload variations \cite{thuruthel_model-based_2019, bruder_koopman-based_2021} and inertial dynamics \cite{haggerty_control_2023}. However, tip-based control leaves the infinite-dimensional body deformation undetermined. This is inadequate for tasks requiring explicit shape control, e.g., navigating through clutter environment.


\begin{figure}[t!]
    \centering
    \includegraphics[width=1.0\linewidth]{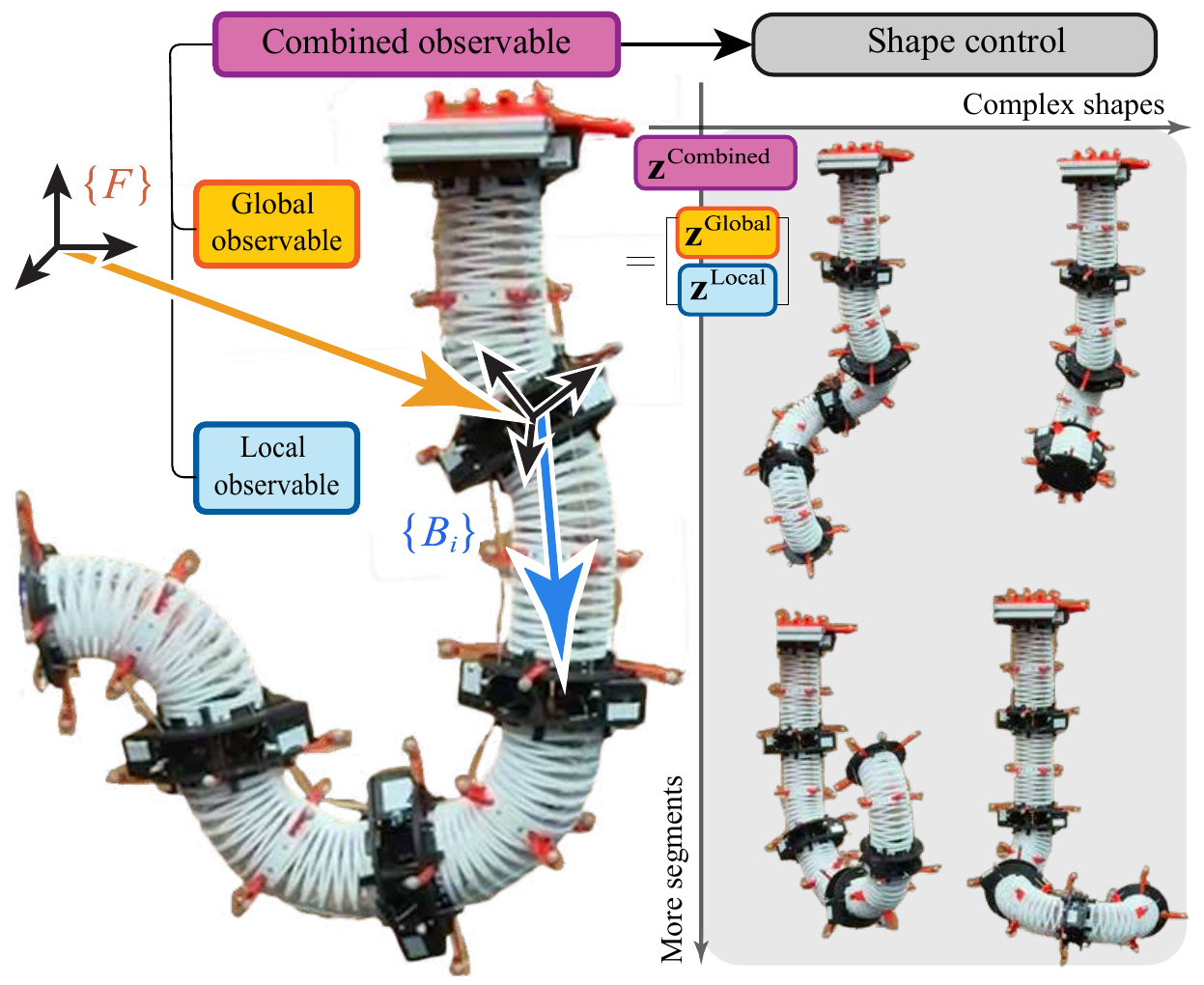}
    \caption{Illustrations for multi-segment soft robot shape control. Global observable: states are expressed in the global fixed frame $\left\{ F \right\} $. Local observable: states are expressed in the segment local frame $\left\{ B_i \right\}$. Combined observable: the proposed global–local observable simultaneously enforces global position and local deformation. 
    }
    \label{fig:schematic}
\end{figure}

Achieving shape control in multi-segment soft robotic arms is significantly more challenging than tip control. As the number of segments increases, each proximal segment must support and compensate for the weight and motion of all distal segments, and this loading changes with the current shape of the robot. Therefore, even without an externally attached payload, multi-segment shape control involves significant gravity-induced coupling between segments. This makes the control problem substantially different from unconstrained single-segment bending or tip-only tracking \cite{ fischer_dynamic_2023, pei_imu_2024, an_shape_2024}.

In practice, closed-loop shape control of multi-segment systems in real-time remains  an open and unresolved challenge. Existing shape control tasks are often restricted to planar deformation \cite{della_santina_model-based_2020,fischer_dynamic_2023}, 
shapes with small deformation \cite{junfeng_shape_2023, tang_general_2026, wang_rl-based_2025, shen_online_2024}, and robots with a limited number of segments, as shown in Tab.~\ref{tab:soft_robot_comparison_caps_onecol}. These limitations arise from challenges in 1) defining appropriate shape metrics, 2) modeling strongly nonlinear continuum dynamics, and 3) achieving computationally efficient real-time control. 


To address these challenges, substantial efforts have been devoted to physics-based methods of soft robotic arms. Continuum-mechanics formulations, such as Cosserat \cite{rucker_statics_2011, till_real-time_2019, renda_dynamic_2014} and Kirchhoff rod models \cite{bao_kinematics_2019}, yield accurate but highly nonlinear dynamical systems governed by partial differential equations (PDEs), which are computationally demanding and difficult to deploy in real-time control. Finite element methods have also been explored for deformation modeling \cite{duriez_control_2013}, yet the spatial and temporal resolution required to capture dynamic behavior limits their practical applicability. Reduced-order and center-of-gravity-based models improve computational efficiency \cite{goury_fast_2018, wang_dynamic_2021}, but often require solving nonlinear optimization problems whose complexity increases with the number of segments \cite{george_thuruthel_control_2018}. As a result, these approaches are typically demonstrated in simulation without real-time shape control. The piecewise constant curvature (PCC) model remains widely adopted due to its simplicity \cite{della_santina_improved_2020, katzschmann_dynamic_2019}. However, the constant-curvature assumption becomes inaccurate in multi-segment configurations. For example, loads induced by distal segments introduce shape-dependent forces on proximal segments, especially for the base segment. Consequently, despite significant progress, existing physics-based approaches struggle with high-dimensional, real-time, multi-segment shape control.

As an alternative, data-driven methods have been explored to learn soft robotic arm dynamics directly from data. Neural networks and recurrent models have been used to capture soft robot dynamics and enable closed-loop tip tracking under disturbances and unmodeled effects \cite{thuruthel_model-based_2019, chen_versatile_2025}. Reinforcement learning and online learning strategies further enable adaptive control under slowly varying conditions, achieving static or quasi-static shape regulation with limited deformation on two segments \cite{shen_online_2024, wang_rl-based_2025}. Vision-based learning approaches have also demonstrated shape matching using image-based observables, typically restricted to planar or low-dimensional deformation \cite{almanzor_static_2023, junfeng_shape_2023}. Despite these advances, learning highly nonlinear, high-dimensional continuum dynamics for multi-segment soft robots remains challenging, and the resulting controllers are often difficult to interpret or integrate into control frameworks due to their black-box nature.

Unlike purely black-box models, Koopman operator-based methods enable interpretable linear predictors and linear controllers \cite{korda_linear_2018, shi_koopman_2025}. However, most existing Koopman-based controllers construct the lifted state from position coordinates along the backbone expressed in the global frame \cite{bruder_modeling_2019,bruder_koopman-based_2021, bruder_data-driven_2021,bruder_koopman-based_2024, haggerty_control_2023,singh_controlling_2023} (we refer to as global observables). These representations are effective for low-dimensional tasks such as tip tracking, but become insufficient for shape control of multi-segment soft robots, which suffer from strongly nonlinear continuum dynamics (segment coupling, gravity, and inertia).
Since global observables only encode pose in the global frame and do not describe deformation in the local frame (which we refer to as local observables), the learned Koopman model will underfit local deformation along the backbone. This motivates the need for a structured observable that captures both global and local information of the multi-segment soft robotic arm.

\begin{table}[t!]
\centering
\caption{Capabilities of multi-segment soft robotic arm closed-loop shape control.}
\label{tab:soft_robot_comparison_caps_onecol}
\small
\setlength{\tabcolsep}{4pt}
\renewcommand{\arraystretch}{1.12}

\begin{tabularx}{\columnwidth}{%
  >{\centering\arraybackslash}p{0.65cm}  
  >{\centering\arraybackslash}p{0.45cm}  
  >{\centering\arraybackslash}p{0.70cm}  
  >{\centering\arraybackslash}p{0.70cm}  
  Y Y Y                                  
}
\toprule
\multirow{2}{*}[-0.3em]{\textbf{Ref.}} &
\multirow{2}{*}[-0.3em]{\textbf{Seg}} &
\multirow{2}{*}[-0.3em]{\textbf{L (m)}} &
\multirow{2}{*}[-0.3em]{\textbf{Model}} &
\multicolumn{3}{c}{\textbf{Shape Control}} \\
\cmidrule(lr){5-7}
& & & &
Planar & 3D Pattern & Continuous tracking \\
\midrule

\cite{almanzor_static_2023}       & 2 & 0.12 & D & \CheckmarkBold & \NA            & \NA \\
\cite{wang_rl-based_2025}         & 2 & 0.33 & D & \NA & \CheckmarkBold    & \NA \\
\cite{junfeng_shape_2023}         & 2 & 0.41 & P & \NA & \CheckmarkBold            & \NA \\
\cite{shen_online_2024}           & 2 & 0.50 & D & \CheckmarkBold & \NA            & \CheckmarkBold \\
\midrule


\cite{kasaei_shape-aware_2025}    & 3 & 0.15 & D & \NA         & \CircleOpen    & \NA \\
\cite{adibnazari_dynamic_2025}    & 3 & 1.12 & P & \NA         & \NA            & \CircleOpen \\
\midrule

\cite{della_santina_model-based_2020} & 6  & 1.00 & P & \CheckmarkBold & \NA         & \CheckmarkBold \\
\cite{tang_general_2026}              & 11 & 0.44 & D & \CheckmarkBold & \NA & \NA \\
\midrule

\textbf{This} & \textbf{3} & \textbf{0.60} & \textbf{D} 
& \CheckmarkBold & \CheckmarkBold & \CheckmarkBold \\
\textbf{work} & \textbf{5} & \textbf{1.00} & \textbf{D} 
& \CheckmarkBold & \CheckmarkBold & \CheckmarkBold \\
\bottomrule
\end{tabularx}

\vspace{1pt}
\footnotesize
\raggedright
\textit{Legend:}
\CheckmarkBold~achieved;
\CircleOpen~ simulation-only result;
---~not addressed or not reported.
L represents the total length of the soft robotic arm.
P and D denote physics-based and data-driven models, respectively.
\end{table}

In this work, we address shape control of multi-segment soft robotic arms by combining global and local observables within a Koopman-based model predictive control (MPC) framework. As illustrated in Fig.~\ref{fig:schematic}, this combined representation (purple) overcomes the drawback in global observables (orange) by adding local observables (blue) in the Koopman lifting function, without changing the underlying model structure. This combined representation captures both global position and local deformation, leading to a better-conditioned control objective in MPC. As a result, the controller can simultaneously reduce global and local shape error, enabling accurate real-time closed-loop tracking of full-body shapes in multi-segment soft robotic systems.

The main contributions of this paper are:

    1) A shape-control framework for multi-segment soft robotic arms. Instead of controlling the end-effector pose alone, the proposed framework regulates the full body configuration. By combining global and local observables within a Koopman-based dense MPC formulation, the controller reduces both global and local shape errors during shape control.
   
    2) Scalability beyond two- or three-segment systems. The proposed framework is not limited to one- or two-segment robots. It is designed for redundant multi-segment soft arms, where many configurations can produce similar end-effector positions. In simulation, the method scales to robots with up to 10 independently actuated segments. In hardware, the validation is performed on 3- and 5-segment robots. 
    
    3) Experimental validation beyond planar or static shape tracking.
    We demonstrate continuous real-time full-body shape tracking on 3- and 5-segment soft robots, including fast tracking, robustness to distal payloads and external disturbances, and a confined-space inspection task. These results show that the proposed framework has the potential to serve as a real-time shape-tracking layer for future multi-segment soft robot systems operating under dynamic motion, unmodeled perturbations, and constrained environments.

    

The remainder of this paper is organized as follows. 
Section~\ref{sec:math_prelims} reviews the mathematical background of Koopman operator theory and the dense MPC formulation. 
Section~\ref{sec:observables} introduces the proposed combined observables, including the lifting function design and shape error metrics. 
Section~\ref{sec:numerical} presents numerical experiments to evaluate the scalability of the proposed method. 
Section~\ref{sec:physical} reports physical experiments on 3-segment and 5-segment soft robotic arms, demonstrating real-time shape control, evaluating robustness, and validating in an inspection task.
Section~\ref{sec:discussion} discusses key insights, including observable design, shape error metric, and limitations. 
Finally, Section~\ref{sec:conclusion} concludes the paper.

\section{Mathematical Background}
\label{sec:math_prelims}

In this section, we briefly review Koopman operator theory and methods for identifying discrete-time Koopman models. We then introduce the dense MPC framework to achieve high-frequency real-time control, which is necessary for such a high-dimensional system. In particular, we highlight a key challenge in practice: the choice of lifting functions directly determines how well such a linear representation can be approximated. This framework provides the theoretical foundation for the lifting function design and control strategy developed in the following sections.

\subsection{Koopman Operator Theory}
\label{sec:koopman_theory}

Consider a nonlinear dynamical system whose time evolution is described by the PDE
\begin{equation}
    \dot{\mathbf{x}} = \mathbf{f}(\mathbf{x}, \mathbf{u}),
\end{equation}
where $\mathbf{x} \in \mathcal{X}$ denotes the system state, $\mathbf{u} \in \mathcal{U}$ represents the control input, and $\mathbf{f}: \mathcal{X} \times \mathcal{U} \rightarrow \mathcal{X}$ defines the system dynamics. In general, $\mathcal{X}$ and $\mathcal{U}$ may lie on nonlinear manifolds. For simplicity, we assume $\mathcal{X} \equiv \mathbb{R}^n$ and $\mathcal{U} \equiv \mathbb{R}^m$.

In discrete time, the system is propagated forward by the flow map $\mathbf{F}_{\Delta t}: \mathcal{X} \times \mathcal{U} \rightarrow \mathcal{X}$ defined over an interval $\Delta t$ by
\begin{equation}
\begin{split}
    \mathbf{x}(t + \Delta t) &= \mathbf{F}_{\Delta t}(\mathbf{x}(t), \mathbf{u}(t)) \\
    &= \mathbf{x}(t) + \int_{t}^{t + \Delta t} \mathbf{f}(\mathbf{x}(\tau), \mathbf{u}(\tau)) ~\di{\tau}.
\end{split}
\end{equation}

To enable the use of linear control techniques, it is common to assume that control inputs are constant over discrete time intervals $\Delta t$, a discretization known as a zero-order hold \cite{proctor_generalizing_2018, bruder_data-driven_2021}. Under this assumption, the dynamics of control inputs $\mathbf{u}$ can be ignored, and the flow map $\mathbf{F}_{\Delta t}$ induces the equivalent discrete time dynamical system
\begin{equation}
\label{eq:discrete_flow_map}
    \mathbf{x}_{k+1} = \mathbf{F}_{\Delta t} (\mathbf{x}_k, \mathbf{u}_k).
\end{equation}


The discrete-time Koopman operator $\mathcal{K}_{\Delta t}$ is an infinite-dimensional linear operator, belonging to an infinite-dimensional Hilbert space $\mathcal{H}$. It acts on scalar-valued function $g: \mathcal{X} \times \mathcal{U} \rightarrow \mathbb{R}$ by propagating them forward in time with the flow map
\begin{equation}
    \mathcal{K}_{\Delta t} g \triangleq g \circ \mathbf{F}_{\Delta t},
\end{equation}
where $\circ$ denotes function composition. However, it is infeasible to compute the full infinite-dimensional Koopman operator. Since a Koopman-invariant subspace can be constructed by the span of a finite set of eigenfunctions of the Koopman operator, one can construct a globally linear representation of the nonlinear system using a finite set of functions \cite{kaiser_data-driven_2021}. To identify these functions, several existing techniques, such as Extended Dynamic Mode Decomposition (EDMD) \cite{williams_datadriven_2015}, enable the approximation of Koopman eigenfunctions and the associated finite-dimensional Koopman operators.

\subsection{Discrete-time Koopman Operator Approximation}
\label{subsec:koopman_approx}

In order to approximate Koopman Operator, Extended Dynamic Mode Decomposition with Control (EDMDc) \cite{proctor_generalizing_2018} is employed as an extension of EDMD for systems with control inputs.
Under the zero-order hold discretization described in Eq.~\eqref{eq:discrete_flow_map}, dataset $D=\{\mathbf{x}_k, \mathbf{x}_{k+1}, \mathbf{u}_k\}_{k=1}^{N-1}$ collects sampled discrete-time trajectories. Let $\boldsymbol{\theta}(\mathbf{x}, \mathbf{u}): \mathcal{X} \times \mathcal{U} \rightarrow \mathbb{R}^M$ denote a vector-valued functions, having $M \gg n + m$. Using this dictionary, we construct the data matrices $\Theta_{X,U}$ and $\Theta_{X^\prime,U}$ such that
\begin{equation} \label{eq:theta_matrix}
    \Theta_{X,U} = \begin{bmatrix}
        \vertbar & & \vertbar \\
        \boldsymbol{\theta}(\mathbf{x}_1, \mathbf{u}_1) & \hdots & \boldsymbol{\theta}(\mathbf{x}_{N-1}, \mathbf{u}_{N-1}) \\
        \vertbar & & \vertbar
    \end{bmatrix},
\end{equation}
and $\Theta_{X^\prime ,U}$ is similarly defined, expect applied to the lifted states after time shift $\boldsymbol{\theta}(\mathbf{x}_{i+1},\mathbf{u}_{i+1},)$, where \(\Theta \in \mathbb{R}^{M \times N} \), and each column corresponds to a lifted state. $X = [\mathbf{x}_0, \ldots, \mathbf{x}_{N-1}]$, 
$U = [\mathbf{u}_0, \ldots, \mathbf{u}_{N-1}]$, 
and $X' = [\mathbf{x}_1, \ldots, \mathbf{x}_N]$ 
denote the state, input, and next-step state matrices.

A finite-dimensional approximation of the discrete-time Koopman operator, denoted $K_{\Delta t}$, can then be obtained by solving the least-squares problem
\begin{align} \label{eq:discrete-time-koop}
\begin{split}
    K_{\Delta t}^* 
    &:= \argmin_{K_{\Delta t}} \left\|K_{\Delta t} \Theta_{X, U} - \Theta_{X^\prime, U} \right\|_2^2 \\
    &\approx \Theta_{X^\prime, U} \Theta^\dagger_{X, U},
\end{split}
\end{align}
where $(\cdot)^\dagger$ denotes the Moore–Penrose pseudoinverse. 
We follow \cite{bruder_modeling_2019, bruder_data-driven_2021} and directly use $K_{\Delta t}$ to propagate the dictionary of functions $\boldsymbol{\theta}$ forward in time, as
\begin{equation} \label{eq:update_rule}
    \boldsymbol{\theta}(\mathbf{x}_{k+1}, \mathbf{u}_k) \approx K_{\Delta t} \boldsymbol{\theta}(\mathbf{x}_{k}, \mathbf{u}_k).
\end{equation}

Computing the pseudo-inverse in Eq.~\eqref{eq:discrete-time-koop} may lead to overfitting and sensitivity to noise or outlier data \cite{seheult_robust_1989}. To mitigate this issue, we use the Least Absolute Shrinkage and Selection Operator (LASSO) \cite{tibshirani_regression_1996,bruder_modeling_2019}. This approach improves the sparsity of learned matrices through $L^1$ regularization, reducing eigenvalue magnitudes and improving the stability of the learned system.
We modify Eq.~\eqref{eq:discrete-time-koop} as
\begin{align} \label{eq:regularization}
\begin{split}
    K_{\Delta t}^* & := \argmin_{K_{\Delta t}} \left\|K_{\Delta t} \Theta_{X,U} - \Theta_{X^\prime,U} \right\|_2^2  + \alpha \|K_{\Delta t}\|_1,
\end{split}
\end{align}
where $\alpha$ is a hyperparameter controlling the magnitude of $L^1$ regularization.

\subsection{Koopman MPC and Dense Form}
\label{subsec:dense_mpc}

MPC is a model-based control strategy that computes optimal control inputs over a finite prediction horizon using a system model. For linear systems, MPC can be formulated as a quadratic program (QP), which can be efficiently solved in real time \cite{korda_linear_2018}. To leverage linear MPC within the Koopman framework, the Koopman operator is expressed in a control-affine form by selecting $\boldsymbol{\theta}(\mathbf{x}, \mathbf{u})$ as:
\begin{equation} \label{eq:control_affine_theta}
        \boldsymbol{\theta}(\mathbf{x}, \mathbf{u}) = \begin{bmatrix}
        \boldsymbol{\theta}_{\mathbf{x}}(\mathbf{x}) \\ \mathbf{u}
    \end{bmatrix},
\end{equation}
where $\boldsymbol{\theta}_{\mathbf{x}}: \mathcal{X} \rightarrow \mathbb{R}^{M-m}$ is state-dependent lifting function. Due to the zero-order hold discretization of $\mathbf{u}$, the Koopman model reduces to linear systems form, using the dictionary given by Eq.~\eqref{eq:control_affine_theta} 
\begin{align} \label{eq:control_affine_form}
\begin{split}
    \boldsymbol{\theta}_{\mathbf{x}}(\mathbf{x}_{k+1}) &= \begin{bmatrix}
        I_{M-m} & 0_{(M-m) \times m}
    \end{bmatrix} K_{\Delta t} \begin{bmatrix}
        \boldsymbol{\theta}_{\mathbf{x}}(\mathbf{x}_k) \\ \mathbf{u}_k
    \end{bmatrix} \\
    &= A \boldsymbol{\theta}_{\mathbf{x}}(\mathbf{x}_k) + B \mathbf{u}_k,
\end{split}
\end{align}
where $A \in \mathbb{R}^{(M-m) \times (M-m)}$, $B \in \mathbb{R}^{(M-m) \times m}$, and 
\begin{equation} \label{eq:k_a_b_determiners}
    K_{\Delta t} = \begin{bmatrix}
        A & B \\
        0_{m \times (M-m)} & I_{m}
    \end{bmatrix}.
\end{equation}

This formulation yields a linear dynamical system in the lifted state $\mathbf{z}_k := \boldsymbol{\theta}_{\mathbf{x}}(\mathbf{x}_k)$ with the original control inputs. 
To recover the original state, we define an output matrix $C$ such that $\mathbf{x}_k = C \mathbf{z}_k$. This is ensured by including the state $\mathbf{x}$ in the dictionary, such that 
\begin{equation} \label{eq:k_c_determiner}
C =
\begin{bmatrix}
I_n & 0_{n\times(M-m-n)}
\end{bmatrix}, \quad
\boldsymbol{\theta}_{\mathbf{x}}(\mathbf{x}) =
\begin{bmatrix}
\mathbf{x} \\
\boldsymbol{\theta}_{\mathbf{x}}^\prime(\mathbf{x})
\end{bmatrix},
\end{equation}
where $\boldsymbol{\theta}_{\mathbf{x}}^\prime : \mathbb{R}^n \rightarrow \mathbb{R}^{M-m-n}$ is an additional vector valued dictionary of functions. 

Over a finite horizon $N_H$, a reference trajectory  $\{\mathbf{r}_k\}_{k=0}^{N_H} \subset \mathcal{X}$ is defined for the system in Eq.~\eqref{eq:control_affine_form}. The tracking error is given by $\mathbf{e}_k = C \mathbf{z}_k - \mathbf{r}_k$. The linear MPC controller determines the control inputs $\{\mathbf{u}_k\}_{k=0}^{N_H-1}$ that minimize the quadratic objective
\begin{align} \label{eq:mpc_error}
\begin{split}
J(\mathbf{z}_0, \bar{\mathbf{u}})
= \sum_{k=0}^{N_H-1} \left( \mathbf{e}_k^\top Q \mathbf{e}_k + \mathbf{u}_k^\top R \mathbf{u}_k \right)
+ \mathbf{e}_{N_H}^\top P \mathbf{e}_{N_H},
\end{split}
\end{align}
subject to linear constraints on the control inputs, where $\mathbf{z}_0$ is the known initial condition, $\bar{\mathbf{u}}:=[{\mathbf{u}_0}^{\top},\cdots ,{\mathbf{u}_{N_H-1}}^{\top}]^{\top}$ is the stacked control input vector, and $Q, P \in \mathbb{R}^{n \times n}$ and $R \in \mathbb{R}^{m \times m}$ are positive semidefinite gain matrices. 

In practice, accurate Koopman operator approximations require high-dimensional dictionaries, i.e., $M \gg n$. Na\"ively propagating the lifted state across the prediction horizon would therefore introduce unnecessary computational overhead. Proposed by \cite{korda_linear_2018}, ``dense'' MPC formulation directly optimizes the sequence of control inputs $\{\mathbf{u}_k\}_{k=0}^{N_H-1}$, rewriting \eqref{eq:mpc_error} into
\begin{align} \label{eq:mpc_objective}
J(\mathbf{z}_0,\bar{\mathbf{u}})=\bar{\mathbf{u}}^{\top}H\bar{\mathbf{u}}+(h^{\top}+{\mathbf{z}_0}^{\top}G)\bar{\mathbf{u}},
\end{align}
 with positive semi-definite matrix $H \in \mathbb{R}^{m{N_H} \times m{N_H}}$, vector $h \in \mathbb{R}^{m{N_H}}$, and matrix $G \in \mathbb{R}^{(M-m) \times m{N_H}}$ can be precomputed based on $A$ , $B$ and gain matrices in Eq.~\eqref{eq:mpc_error} \cite{korda_linear_2018}. Consequently, the dense MPC can be solved independently of the large lifting dimension $M$. In this work, the resulting QP is solved using  OSQP solver \cite{stellato_osqp_2020}, enabling the update rate exceeding 300~Hz on a standard CPU (2.2~GHz).

\section{Global, Local, and Combined Observables} \label{sec:observables}

In this section, we review global observables (benchmark) and introduce our method: combined observables (proposed method) by integrating global and local observables.
These observables define the structure of the lifted state and directly influence the effectiveness of the Koopman model and the resulting control performance.

\subsection{Lifting Function Design}

\begin{figure*}
    \centering
    \includegraphics[width=1.0\linewidth]{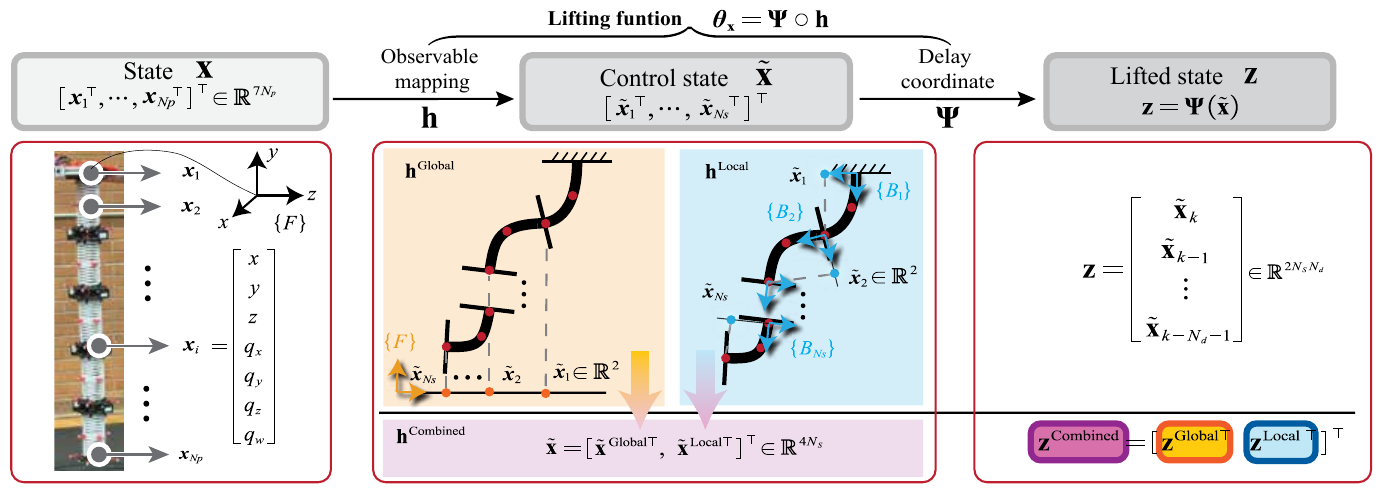}
    \caption{Schematic illustration of global, local, and combined observable mappings and their role in lifting function. The lifting function is defined as $\boldsymbol{\theta}_\mathbf{x} = \mathbf{\Psi} \circ \mathbf{h}$, where delay coordinates $\mathbf{\Psi}$ are applied to obtain the lifted state $\mathbf{z}_k = \mathbf{\Psi}(\tilde{\mathbf{x}}_k)$ for Koopman-based prediction and control. The measured state $\mathbf{x}_k$ is mapped to a reduced control state $\tilde{\mathbf{x}}_k = \mathbf{h}(\mathbf{x}_k)$. Global observable captures task-space positions in global fixed frame $\left\{ F \right\}$, while local observable captures segment-wise deformation in local base frames $\left\{ B_j \right\}$. The combined observable concatenates both representations to retain global consistency and local detail.
}
    \label{fig:obs_mapping}
\end{figure*}

In Koopman-based modeling, the choice of lifting function $\boldsymbol{\theta}_{\mathbf{x}}$ plays a critical role in both model accuracy and control performance. 
Existing Koopman-based method
commonly construct the lifted state $\mathbf{z}_k$ from measured state $\mathbf{x}_k$ augmented with delay coordinates \cite{bruder_modeling_2019,bruder_koopman-based_2021,bruder_data-driven_2021,bruder_koopman-based_2024,haggerty_control_2023,singh_controlling_2023}. 

However, this representation contains geometric information that is not directly relevant to the control objective and may degrade model performance. For example, incorporating orientation requires constructing observables that preserve quaternion normalization or enforce $SO(3)$ constraints, significantly increasing model complexity without directly improving shape control performance. Since our objective is to control backbone geometry shape, we instead define a reduced control state $\tilde{\mathbf{x}}_k \in \mathbb{R}^{\tilde{n}}$ that captures inherent geometric features.
\begin{equation}
\tilde{\mathbf{x}}_k = \mathbf{h}(\mathbf{x}_k),
\end{equation}
where $\mathbf{h}: \mathbb{R}^n \rightarrow \mathbb{R}^{\tilde{n}}$ is the observable mapping that extracts geometric features from the measured state, will be further described in Sec.~\ref{subsec:obs_map_design}.

Since $\tilde{\mathbf{x}}_k$ contains only instantaneous geometric information and does not explicitly encode dynamic effects such as velocities or internal forces, we augment the lifted state using delay coordinates to account for these latent dynamics  \cite{haggerty_control_2023}. Formally, let $\mathbf{\Psi}: \mathbb{R}^{\tilde{n}} \rightarrow \mathbb{R}^{\tilde{n} ({N_d}+1)}$ denote a delay coordinate with horizon $N_d$, the lifting function $\boldsymbol{\theta}_{\mathbf{x}}$ is defined as
\begin{align} \label{eq:delay_coordinate_embedding}
\begin{split}
\mathbf{z}_k &= \boldsymbol{\theta}_{\mathbf{x}}(\mathbf{x}_k) = \mathbf{\Psi}(\mathbf{h}(\mathbf{x}_k)) \\
&= \left[ \tilde{\mathbf{x}}_k^\top,\ \tilde{\mathbf{x}}_{k-1}^\top,\ \cdots,\ \tilde{\mathbf{x}}_{k-N_d}^\top \right]^\top.
\end{split}
\end{align}

As illustrated in Fig.~\ref{fig:obs_mapping}, the lifting function is constructed as the composition $\boldsymbol{\theta}_{\mathbf{x}} = \mathbf{\Psi} \circ \mathbf{h}$, where the observable mapping $\mathbf{h}$ defines the control-relevant state and the delay embedding $\mathbf{\Psi}$ captures system dynamics. This formulation highlights that the design of $\mathbf{h}$ is central to the Koopman representation and directly impacts control performance, which motivates the observable mappings introduced in the following subsection.

\subsection{Observable Mapping Design}
\label{subsec:obs_map_design}

In this paper, we consider three types of observable mappings: $\mathbf{h}^{\mathrm{Global}}$ (benchmark), $\mathbf{h}^{\mathrm{Local}}$, and $\mathbf{h}^{\mathrm{Combined}}$ (proposed method) as shown in the middle box of Fig.~\ref{fig:obs_mapping}. We discretize the backbone into $N_p$ points. Each point defines a frame $\{B_i\} \in SE(3)$, characterized by position $\mathbf{p}_i \in \mathbb{R}^3$ and orientation $\mathbf{q}_i \in \mathbb{R}^4$, where $\mathbf{q}_i$ is the unit quaternion corresponding to a rotation matrix $R_i \in SO(3)$. The full measured state is $\mathbf{x} = [\mathbf{p}_1{^\top}, \mathbf{q}_1{^\top}, \dots, \mathbf{p}_{N_p}{^\top}, \mathbf{q}_{N_p}{^\top}]^\top$, expressed in global frame $\left\{ F \right\} $. Here, $\{B_1\}$ corresponds to the base frame and $\{B_{N_p}\}$ to the tip frame. We denote $\mathbf{p}_i^{\{B_j\}}$ as the position of the $i$th point expressed in frame $\{B_j\}$. The global frame $\{F\}$ is defined to coincide with the base frame $\{B_1\}$.

\subsubsection{Global observable}
\label{subsubsec:global_obs}
A common approach is to choose state as positions $\mathbf{p}_i^{\{F\}}$ and orientations $\mathbf{q}_i^{\{F\}}$ in the global frame  \cite{chen_versatile_2025,bruder_koopman-based_2024, adibnazari_dynamic_2025, pei_imu_2024}. This approach does not use observable mapping. It works well for low-dimensional problems such as tip control tasks or plan shape control with a limited number of segments ($\leqslant 3$).

However, using the full coordinates of  $\mathbf{p}_i^{\{F\}}$ leads to instability in shape control, potentially due to the large modeling error in the extension direction of the soft robot, arising from its high nonlinearity.
To address this issue, a reduced global observable was constructed, projecting positions onto the $x$-$z$ plane (normal plane) of the global frame $\{F\}$, effectively removing the poorly modeled extension direction \cite{wang_dataefficient_2026}. This simplification improves model consistency and results in more stable control behavior. 

Formally, let $\bar{R}^{\{F\}} \in \mathbb{R}^{2 \times 3}$ denote the projection matrix onto the $x$--$z$ plane of $\{F\}$. The global observable is defined as
\begin{equation}
\mathbf{h}^{\mathrm{Global}}(\mathbf{x})=\left[ (\bar{R}^{\{F\}}\mathbf{p}_1)^{\top},\cdots ,(\bar{R}^{\{F\}}\mathbf{p}_{N_p})^{\top} \right] ^{\top}.
\end{equation}

After delay coordinate embedding $\mathbf{\Psi}$ in Eq.~\eqref{eq:delay_coordinate_embedding} to obtain lifted state $\mathbf{z}_k^{\text{Global}}$, the final Koopman model takes the form
\begin{align}
\begin{split}
    \mathbf{z}_{k+1}^{\text{Global}} &= A^{\text{Global}} \mathbf{z}_{k}^{\text{Global}} + B^{\text{Global}} \mathbf{u}_k \\
    \tilde{\mathbf{x}}_k &= C^{\text{Global}} \mathbf{z}_k,
\end{split}
\end{align}
where $A^{\text{Global}}$ and $B^{\text{Global}}$ are defined in Eq.~\eqref{eq:k_a_b_determiners}, and $C^{\text{Global}}$ follows Eq.~\eqref{eq:k_c_determiner}.

\subsubsection{Local observable} 
\label{subsubsec:local_obs}

Distinct from the global observable, we introduce segment-frame information as the local observable for Koopman modeling, which becomes increasingly important for multi-segment soft robotic arms. 

Let $N_s$ denote the number of segments in the soft robotic arm. Each measured
point belongs to one specific segment, and we define a segment-assignment map
$\lambda(i)$ that returns the segment containing the $i$th point. For each segment, a local frame $\{B_{\lambda(i)}\}$ is defined for the base segment. The position of the $i$th point expressed in its corresponding local frame is denoted by $\mathbf{p}_i^{\{B_{\lambda(i)}\}}$. Same as global observables, we project these positions onto the normal plane of each segment frame using projection matrices $\bar{R}^{\{B_{\lambda(i)}\}} \in \mathbb{R}^{2 \times 3}$. The local observable is then defined as

\begin{equation}
\mathbf{h}^{\mathrm{Local}}(\mathbf{x}) =
\left[
(\bar{R}^{\{B_{\lambda(1)}\}} \mathbf{p}_1)^\top,\,
\cdots,\,
(\bar{R}^{\{B_{\lambda(N_p)}\}} \mathbf{p}_{N_p})^\top
\right]^\top.
\end{equation}

Similarly, the lifted state under local observable is $\mathbf{z}_k^{\text{Local}} := \boldsymbol{\theta}_{\mathbf{x}}(\mathbf{x}_k)
= \mathbf{\Psi}(\mathbf{h}^{\text{Local}}(\mathbf{x}_k)).$ The resulting Koopman model is
\begin{align}
\begin{split}
    \mathbf{z}_{k+1}^{\text{Local}} &= A^{\text{Local}} \mathbf{z}_{k}^{\text{Local}} + B^{\text{Local}} \mathbf{u}_k \\
    \tilde{\mathbf{x}}_k &= C^{\text{Local}} \mathbf{z}_k,
\end{split}
\end{align}
where $A^{\text{Local}}$ and $B^{\text{Local}}$ are determined by Eq.~\eqref{eq:k_a_b_determiners}, and $C^{\text{Local}}$ follows Eq.~\eqref{eq:k_c_determiner}.

The introduced local observable better captures segment-wise deformation by expressing each point in its corresponding segment frame. However, it does not explicitly encode the task-space position. Small local errors can propagate along the serial chain, leading to global drift (will be shown in Fig.~\ref{fig:error_compare}). This motivates the combined observable introduced next.

\subsubsection{Combined observable}
To exploit the complementary strengths of the global and local observables, we further propose combining them into  a unified representation
\begin{align}
\mathbf{h}^{\mathrm{Combined}}(\mathbf{x})=\left[ \mathbf{h}^{\mathrm{Global}}(\mathbf{x})^{\top},\mathbf{h}^{\mathrm{Local}}(\mathbf{x})^{\top} \right] ^{\top}.
\end{align}

Rather than learning a new Koopman model from the combined observable, we find that it is more accurate to simply concatenate the lifted states $\mathbf{z}_{k}^{\text{Global}}$ and $\mathbf{z}_{k}^{\text{Local}}$. Specifically, the combined lifted state is defined as
\begin{align} 
    \mathbf{z}_{k}^{\text{Combined}}=\left[ {\mathbf{z}_{k}^{\text{Global}}}^{\top},{\mathbf{z}_{k}^{\text{Local}}}^{\top} \right] ^{\top}.
\end{align}

The corresponding Koopman model can then be constructed directly from the independently trained global and local models (Sec.~\ref{subsubsec:global_obs} and ~\ref{subsubsec:local_obs}) as
\begin{align} 
\begin{split}
\mathbf{z}_{k+1}^{\text{Combined}} &= \begin{bmatrix}
	A^{\text{Global}}&		0\\
	0&		A^{\text{Local}}\\
\end{bmatrix} \begin{bmatrix}
	\mathbf{z}_{k}^{\text{Global}}\\
	\mathbf{z}_{k}^{\text{Local}}\\
\end{bmatrix} + \begin{bmatrix}
	B^{\text{Global}} \\		B^{\text{Local}}\\
\end{bmatrix} \mathbf{u}_k \\
\tilde{\mathbf{x}}_k &= \begin{bmatrix}
    C^{\text{Global}} & 0 \\ 0 & C^{\text{Local}}
\end{bmatrix} \mathbf{z}_k^{\text{Combined}}.
\end{split}
\end{align}

Note that, because of the block-diagonal form, the combined Koopman model does not explicitly improve prediction accuracy. Instead, the benefit of the combined observable arises from a richer state representation that improves the conditioning of the MPC objective. 

Although the physical system is inherently coupled, this block-diagonal structure should be interpreted as a modeling approximation in the lifted space rather than a true dynamics decoupling. The coupling between global and local representations is implicitly captured through the observable mapping.

\subsection{Global and Local Shape Error}
\label{subsec:error_metric}

\begin{figure}
    \centering
    \includegraphics[width=0.75\linewidth]{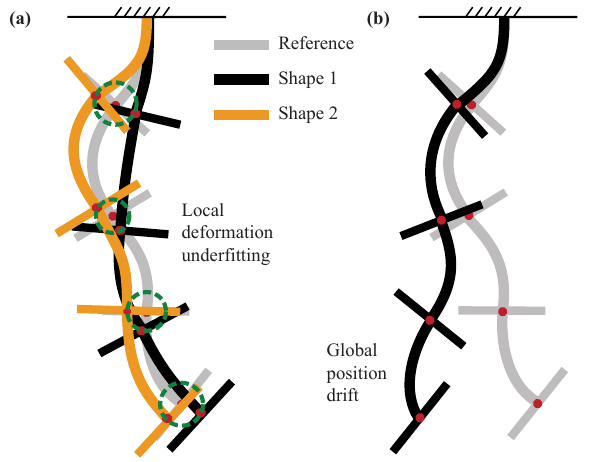}
    \caption{The drawbacks of using global and local shape alone. (a) The same global shape error can correspond to different shapes (local deformation underfitting). (b) For a control that only relies on local shape error, a small local error can accumulate into a large global position drift.
}
    \label{fig:error_compare}
\end{figure}

The global shape error is defined as the mean squared error (MSE) between corresponding backbone points of the actual shape and the reference shape, both expressed in the global frame $\{F\}$. Given $N_p$ measured points with positions $\mathbf{p}_i$, and reference positions $\mathbf{p}_i^\star$, the global shape error is defined as
\begin{align} \label{eq:global_error}
\begin{split}
e^{\text{global}} = \frac{1}{N_p-1} \sum_{i=1}^{N_p} \left\| \mathbf{p}_i - {\mathbf{p}_i^\star} \right\|_2^2,
\end{split}
\end{align}
where the error is averaged over the $N_p-1$ points since the first point ($i=1$) corresponds to the fixed base point. This metric computes the average squared distance error over all deformable backbone points. Notably, the term at $i=N_p$ corresponds to the tip error as a special case. This metric captures the global shape discrepancy in terms of point positions. Some similar definitions could be found in \cite{zhang_stochastic_2025, singh_controlling_2023, tang_general_2026}. While $e^{\text{global}}$ measures the global position error, using it alone can underfit local deformation. Different segment-wise deformation patterns may yield similar global errors, as illustrated in Fig.~\ref{fig:error_compare}~(a).  

To capture local deformation, we define a local shape error by expressing each point in its corresponding segment frame. Specifically,
\begin{equation} 
\label{eq:local_error}
e^{\text{local}} = \frac{1}{N_p-1} \sum_{i=1}^{N_p} \left\| 
\mathbf{p}_i^{\{B_{\lambda(i)}\}} - {\mathbf{p}_i^\star}^{\{B_{\lambda(i)}\}} 
\right\|_2^2,
\end{equation}
where $\lambda(i)$ is defined in Sec.~\ref{subsubsec:local_obs}. However, $e^{\text{local}}$ alone is also insufficient because local errors can accumulate along the serial chain, leading to global task-space drift, as illustrated in Fig.~\ref{fig:error_compare}~(b). Therefore, to fully evaluate shape-control
performance, we report both $e^{\text{global}}$ and $e^{\text{local}}$.


Although the shape error metrics in Eq.~\eqref{eq:global_error} and Eq.~\eqref{eq:local_error} have a quadratic form similar to the MPC objective in Eq.~\eqref{eq:mpc_objective}, the MPC does not directly minimize $e^{\mathrm{global}}$ or $e^{\mathrm{local}}$. This is because the control optimization is performed in the lifted state $\mathbf{z}$, which is constructed from reduced observables and delay embeddings rather than the full geometric state. As a result, the exact geometric error metrics (e.g., orientation) cannot be expressed directly as quadratic functions of $\mathbf{z}$. Instead, the observable mappings and lifting functions are designed such that minimizing the MPC objective indirectly drives both global and local shape errors to decrease. 
In particular, for the combined observable, the MPC cost is defined using a block-diagonal weighting matrix $Q^{\mathrm{Combined}} = \operatorname{diag}(Q^{\mathrm{Global}}, Q^{\mathrm{Local}})$, which jointly penalizes deviations in both global and local representations.
The corresponding terminal cost $P$, control penalty $R$, and output matrix $C$ are defined consistently with this diagonal structure.

\subsection{Special Considerations}

In addition to the regularization introduced in Eq.~\eqref{eq:regularization}, several practical considerations are incorporated.

First, the control inputs in both simulation and physical experiments are defined as input increments. Rapid variations in actuation can induce highly nonlinear inertial dynamics, which are difficult to be captured by Koopman models. To ensure that the system operates within a well-behaved dynamical regime, we formulate the MPC problem in terms of input increments $\Delta \mathbf{u}$ rather than absolute inputs $\mathbf{u}$. Given an initial input $\mathbf{u}_0$, this formulation is equivalent but allows explicit control over the change rate of actuation. In addition, by constraining $\Delta \mathbf{u}$, we prevent the system from entering regimes dominated by unmodeled inertial effects while still enabling sufficient actuation for meaningful motion. Specifically, control inputs are subject to $\|\Delta \mathbf{u}_k\|_\infty \le \Delta u_{\max}$.

Second, in this work, reference trajectories $\mathbf{r}_k$ are constructed from experimentally collected motions generated by predefined input sequences that are excluded from the Koopman training dataset. This is mainly because reference trajectory generation is not the focus of this work, although several shape planning methods exist \cite{veil_shape-space_2026, liu_path_2023, mbakop_parametric_2022}.

Finally, the control state only includes a reduced state representation for the tip points of each segment. In our physical experiments, adding additional intermediate points does not improve prediction accuracy and instead reduces control stability under dense measurements. For consistency, the same representation is used in the simulation. However, the numerical experiment results do not exhibit this instability due to the absence of measurement noise and hardware effects.
\section{Numerical Experiments}
\label{sec:numerical}

In this section, we validate the proposed method and evaluate its scalability with respect to the number of segments using high-fidelity simulations. Due to the high accuracy of the state observation, error accumulation effects are significantly reduced in the simulation. As a result, the performance difference between local and combined observables is less pronounced compared to physical experiments (in Sec. \ref{sec:physical}). Therefore, we focus on comparing global (benchmark) and combined (proposed method) observables for the shape control of multi-segment soft robotic arms. The objective is to assess how each method scales as the number of segments increases, and to demonstrate that the proposed method enables consistent and stable tracking in higher-dimensional systems.

\begin{figure}[t]
    \centering
    \includegraphics[width=0.9\linewidth]{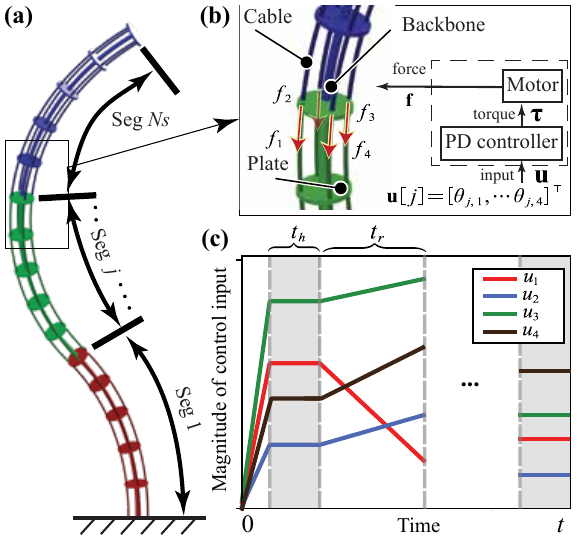}
    \caption{Simulation and data collection setup. 
    (a) Multi-segment soft robotic arm model, showing the flexible backbone, segmented structure, and uniform cable routing. 
    (b) Cable-driven actuation mechanism for each segment. Control inputs $\mathbf{u}_j = [\theta_{j,1}, \dots, \theta_{j,4}]^\top$ are motor angles regulated by a PD controller, which generate motor torques and corresponding cable forces applied through the routing plates.
    (c) Control input generation using a ramp-and-hold scheme, where randomly sampled inputs are linearly interpolated over a ramp phase $t_r$ and a hold phase $t_h$.}
    \label{fig:SimulationSetup}
\end{figure}

\subsection{Description of Simulation Setup}
\label{subsec:numerical_setup}

To simulate the soft robotic arm, the system is separated into three main components: the backbone, the cables, and the plates, as shown in Fig.~\ref{fig:SimulationSetup}. The backbone is modeled using the Kirchhoff rod formulation, subject to cable forces as described by \citet{rucker_statics_2011}. The resulting dynamics are solved using the implicit BDF-$\alpha$ method \cite{till_real-time_2019}. Under this discretization, the governing PDEs are
\begin{align} \label{eq:kirchhoff_model}
\begin{split}
    \mathbf{p}_t &= R \mathbf{q} \\
    R_t &= R \hat{\boldsymbol{\omega}} \\
    \mathbf{v}_t &= \boldsymbol{\omega}_s + \hat{\mathbf{v}} \boldsymbol{\omega} \\
    \mathbf{q}_t &= \frac{1}{\rho A} R^\intercal \mathbf{f} - \hat{\boldsymbol{\omega}} \mathbf{q} \\
    \boldsymbol{\omega}_t &= (\rho J)^{-1} \left(
        K_{bt} (\mathbf{v}_s - \mathbf{v}_s^*) + \hat{\mathbf{v}} K_{bt} (\mathbf{v} - \mathbf{v}^*) \right.\\
        &\qquad + \left. R^\intercal \mathbf{l} - \hat{\boldsymbol{\omega}} \rho J \boldsymbol{\omega}
    \right)
\end{split}
\end{align}
where $\mathbf{p}$ is the backbone position along arclength $s$, $R$ is the orientation, $\mathbf{v}$ is the angular strain, $\mathbf{q}$ is the linear velocity, and $\boldsymbol{\omega}$ is the angular velocity. The subscripts $(\cdot)_s$ and $(\cdot)_t$ denote partial derivatives with respect to arclength $s$ and time $t$, respectively, and $(\cdot)^*$ denotes the reference configuration. The mapping $(\cdot)^\wedge : \mathbb{R}^3 \to so(3)$ denotes the skew-symmetric operator. The parameters $\rho$, $A$, and $J$ denote the material density, cross-sectional area, and second moment of area, respectively, while $K_{bt}$ is the bending–torsion stiffness matrix. The vectors $\mathbf{f}$ and $\mathbf{l}$ represent external forces and moments due to cable actuation. For more details on the BDF-$\alpha$ method, the evaluation of cable forces, and all relevant boundary conditions, we refer the reader to \citet{till_real-time_2019}. 

To model actuation, we use a PD controller to represent the motors as virtual springs. The cable force is computed from motor torque as $\mathbf{f}=\boldsymbol{\tau }/r_{\text{motor}}$, where $r_{\text{motor}}$ is the motor pulley radius (see Fig. \ref{fig:PhysicalSetup}). The torque is generated from the motor angle $\theta$ as 
\begin{equation}
    \tau := K_p (\theta - \theta_{\mathrm{ref}}) + K_d (\dot{\theta} - \dot{\theta}_{\mathrm{ref}}),
\end{equation}
where $K_p$ and $K_d$ are controller gains, $\theta_{\mathrm{ref}}$ is a reference angle outputted by the higher level MPC controller described in Sec.~\ref{subsec:dense_mpc}. The solutions of Eq.~\eqref{eq:kirchhoff_model} are used as ground truth for control experiments and for generating training trajectories for Koopman model learning. The material and simulation parameters are summarized in Table~\ref{tab:sim_params}.

\begin{table}[t]
\centering
\caption{Material, geometric, actuation, and simulation parameters of the cable-driven soft robot model.}
\begin{tabular}{l l c c}
\toprule
\textbf{Category} & \textbf{Parameter} & \textbf{Unit} & \textbf{Value} \\
\midrule

\multirow{4}{*}{Backbone} 
& Total backbone length & m & 1.0 \\
& Young's modulus & GPa & 200 \\
& Density $\rho$ & kg/m$^3$ & 8000 \\
& Backbone radius & mm & 1 \\

\midrule

\multirow{3}{*}{Cable} 
& Number of cables per segment & -- & 4 \\
& Cable offset & cm & 5 \\
& Cable angles & deg & 0, 90, 180, 270 \\

\midrule

\multirow{3}{*}{Actuation} 
& Motor pulley radius & mm & $\text{5}$ \\
& Proportional gain $K_p$ & -- & $\text{6}$ \\
& Derivative gain $K_d$ & -- & $\text{4}$ \\

\midrule

\multirow{2}{*}{Simulation} 
& Spatial discretization (per segment) & -- & 100 \\
& Time step $\Delta t$ & s & 0.01 \\

\bottomrule
\end{tabular}
\label{tab:sim_params}
\end{table}

\begin{figure*}
    \centering
    \includegraphics[width=1\linewidth]{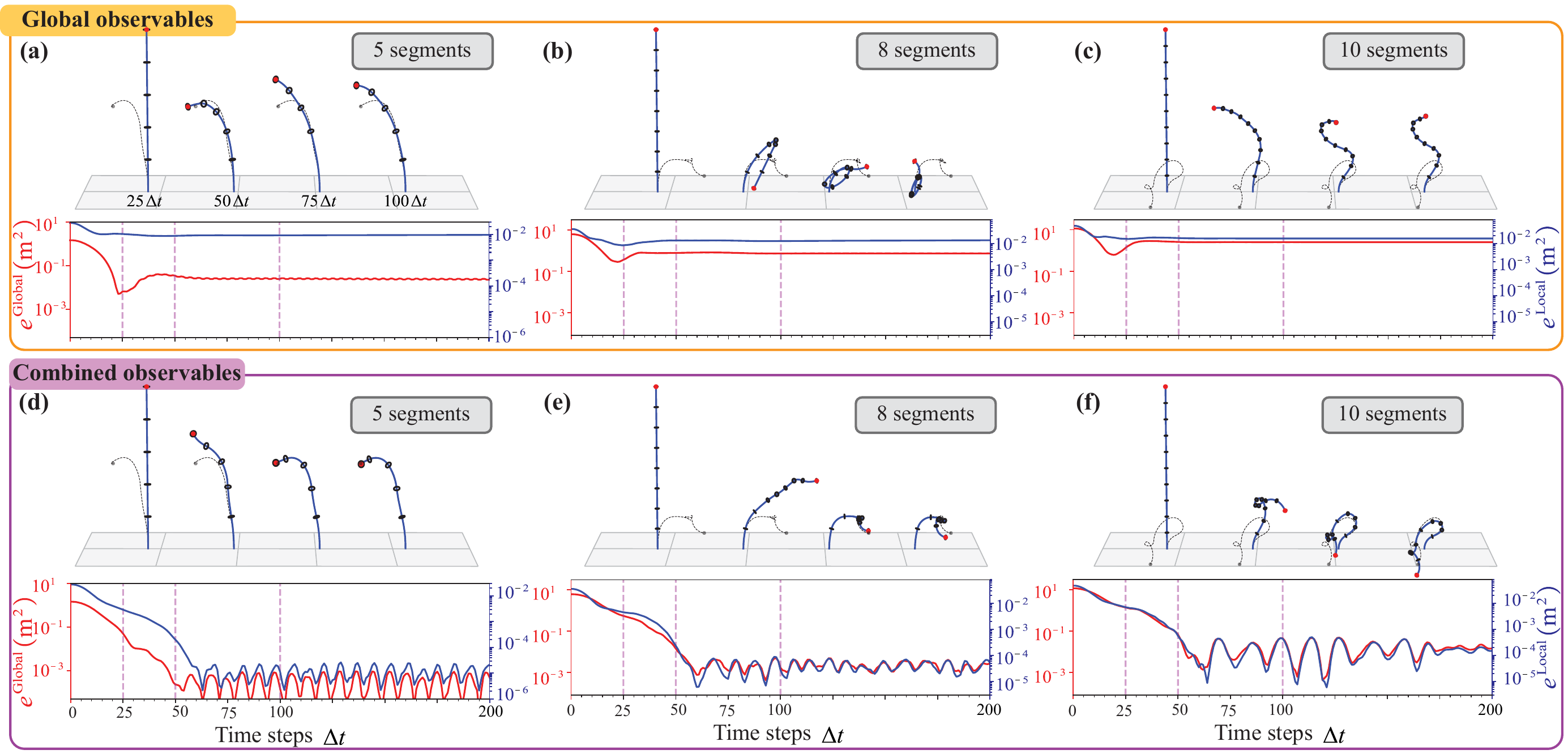}
    \caption{Comparison of global and combined observables for multi-segment soft robotic arm shape control. 
    (a–c) Global observables for 5-, 8-, and 10-segment robots, showing tracking failure as system complexity increases. 
    (d–f) Combined observables for the same systems, demonstrating successful tracking across all segment configurations. 
    In each panel, the dashed black curves denote the reference shapes, and the blue curves represent the robot trajectories at selected time steps. The plots below show global (red) and local (blue) shape errors over time. 
    Note that the results of local observables are similar to the combined case and are omitted for clarity.
    }
    \label{fig:numerical_static_res}
\end{figure*}

For Koopman model construction, the state $\mathbf{x}$ is defined using observations obtained directly from simulation data, consistent with Sec.~\ref{sec:observables}. The control input vector $\mathbf{u} \in \mathbb{R}^m$ is defined as the reference angle of each motor, where $m$ is the total number of motors. In each segment, four cables are routed parallel to the backbone, with opposing cable pairs driven by the same motor, as described in Sec.~\ref{subsec: physical_setup}. Consequently, each segment has two independent control inputs. As described in Sec.~\ref{sec:observables}, the Koopman model is constructed using combined observables with delay embedding $N_d = 3$. For comparison, a baseline model is also trained using only global observables.

For data collection, 200,000 snapshots $(\mathbf{x}(t), \mathbf{x}(t+\Delta t), \mathbf{u}(t))$ are generated for 3-, 5-, 8-, and 10-segment soft robotic arms. The dataset size is selected to ensure convergence of all trained Koopman models. Each simulation is initialized with all motors at rest. Control inputs are generated by sampling $N$ random vectors $\mathbf{u}_i \in \mathbb{R}^m$, where $i = 1, \hdots, N$. Each component is drawn from a uniform distribution $\mathbf{u} \in [-4\pi, 4\pi]$. The input sequence is constructed using a ramp-and-hold scheme, as shown in Fig.~\ref{fig:SimulationSetup}~(c). We allow control inputs to linearly interpolate or ramp between each consecutive $\mathbf{u}_i$ for time $t_r$ before holding them fixed at $\mathbf{u}_i$ for time $t_h$. Specifically, for $t \in [0, N(t_r + t_h)]$,
\begin{equation}
    \mathbf{u}(t) = \begin{cases}
        \mathbf{u}_i (1 - \beta) + \mathbf{u}_{i + 1} \beta & t - t_0 < t_r \\
        \mathbf{u}_{i + 1} & t_r \leq t - t_0 < t_h
    \end{cases},
\end{equation}
where $\mathbf{u}_0 = 0$, and we define $i = \lfloor t / (t_r + t_h) \rfloor$, $t_0 = (t_r + t_h) i$, and $t_1 = (t_r + t_h) (i + 1)$. The interpolation parameter $\beta$ is defined as

\begin{equation}
    \beta = \frac{t - t_0}{t_1 - t_0}.
\end{equation}

We define the Koopman time interval $\Delta t = 0.01$s, ramping and holding times $t_r = 40 \Delta t$ and $t_h = 10 \Delta t$, and choose $N = 4000$. A sample of the control inputs is shown in Fig.~\ref{fig:SimulationSetup}~(b). Snapshots for training the Koopman model are generated by subsampling from these simulations at intervals $\Delta t$. 

\subsection{Numerical Experiment Results}
In the numerical experiments, we specify a static reference shape and evaluate the resulting dynamic trajectories as the soft robot attempts to track this target. The reference shapes are generated by solving the static equilibrium equations derived from Eq.~\eqref{eq:kirchhoff_model} for randomly sampled control inputs $\mathbf{u}$, ensuring that the resulting shapes lie within the training distribution of the Koopman models. For each observable, we evaluate both convergence behavior and steady-state accuracy using the global and local shape errors defined in Eq.~\eqref{eq:global_error} and Eq.~\eqref{eq:local_error}. In particular, we focus on how performance scales with the number of segments, which directly increases system dimensionality and shape complexity.

Fig.~\ref{fig:numerical_static_res} shows representative tracking trajectories and the error curves for 5-, 8-, and 10-segment robots. Under global observables (Fig.~\ref{fig:numerical_static_res}~(a–c)), the Koopman controller fails to accurately track the reference shapes as system complexity increases, converging to configurations with large residual error. In contrast, the combined observables (Fig.~\ref{fig:numerical_static_res}~(d–f)) is able to obtain several orders of magnitude greater accuracy, demonstrating significantly improved scalability. While both methods exhibit transient oscillations due to the underlying system dynamics, the global observable controller converges to a higher steady-state error, whereas the combined observable controller achieves substantially lower error levels. Due to log scaling, although oscillations in the combined observables appear large, they are of the same magnitude as the oscillations in the global controller as well, and arise due to the dynamics of the soft robot.

\begin{figure}[t!]
    \centering
    \includegraphics[width=1\linewidth]{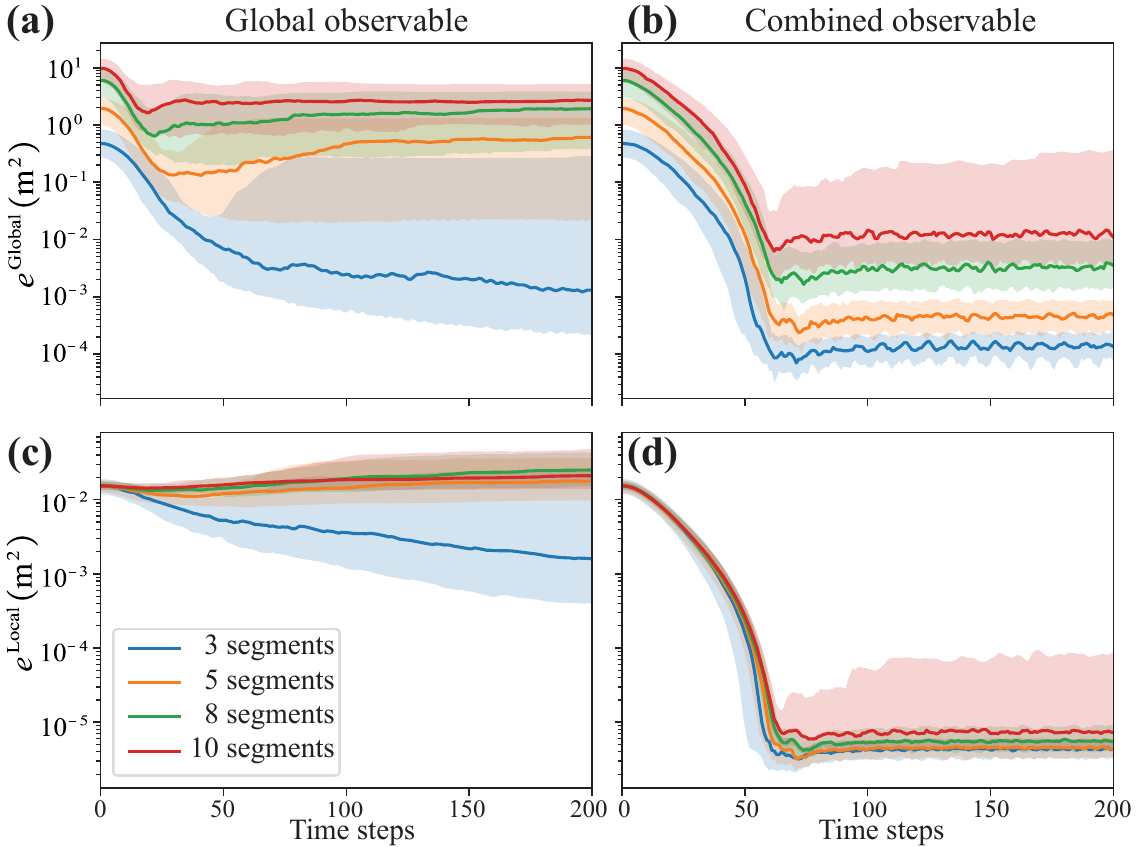}
    \caption{
    Global shape error under (a) global-observable controller and (b) combined-observable controller. 
    Local shape error under (c) global-observable controller and (d) combined-observable controller. 
    Each curve corresponds to a different number of segments (3, 5, 8, and 10).  Shaded regions denote the 25th to 75th percentile range of corresponding errors, and plotted lines correspond to the 50th percentile. Note that the results of local observables very close to the combined case and are omitted for clarity.
    }
    \label{fig:convergence}
\end{figure}
In Fig.~\ref{fig:numerical_static_res}, the global-observable controller exhibits a transient reduction in global shape error around $25$ time steps, followed by convergence to a higher-error configuration. This behavior is caused by steady-state drift arising from modeling inaccuracies in the Koopman approximation. Due to the diagonal format introduced in Sec.~\ref{subsec:obs_map_design},  a similar effect is observed in the combined-observable controller; however, its magnitude is significantly reduced.

Fig.~\ref{fig:convergence} summarizes the statistical performance of the controllers over 200 simulations, each with 200 time steps.  Compared to the global-observable controller, the combined-observable controller has orders of magnitude higher accuracy across all robots in both the global and local error metrics. Further, the combined-observable controller also has orders of magnitude smaller variance as well, visualized by the 25th to 75th percentile ranges in Fig.~\ref{fig:convergence}. As the number of segments increases, all controllers see a decrease in performance in each metric. In particular, the converged global shape error of the combined-observable controller empirically grows exponentially, whereas the global-observable controller fails when there are more than 3 segments.

 Note that, in addition to the displacement-driven simulations presented above, we also evaluated a three-cable force-driven actuation setting with dense measurements along the backbone. In this case, the Koopman controller based on global observables and combined observables failed to converge, while the local observable controller remained stable. This motivates introducing a PD motor controller for the system. These results are omitted due to space limitations, but further support the conclusions regarding observable design.
\section{Physical Experiments}
\label{sec:physical}

In this section, we evaluate the proposed method on two prototypes: a 3-segment robot and a 5-segment robot. We first compare the tracking performance of global, local, and combined observables under identical reference trajectories on a 3-segment robot. Then, we further evaluate the effectiveness of the proposed method on the 5-segment robot using multiple target shapes of various deformation patterns. We also evaluate the controller's robustness under external loads. Finally, a confined-space demonstration shows that the proposed method can serve for future inspection tasks.

\begin{figure}[t!]
    \centering
    \includegraphics[width=1\linewidth]{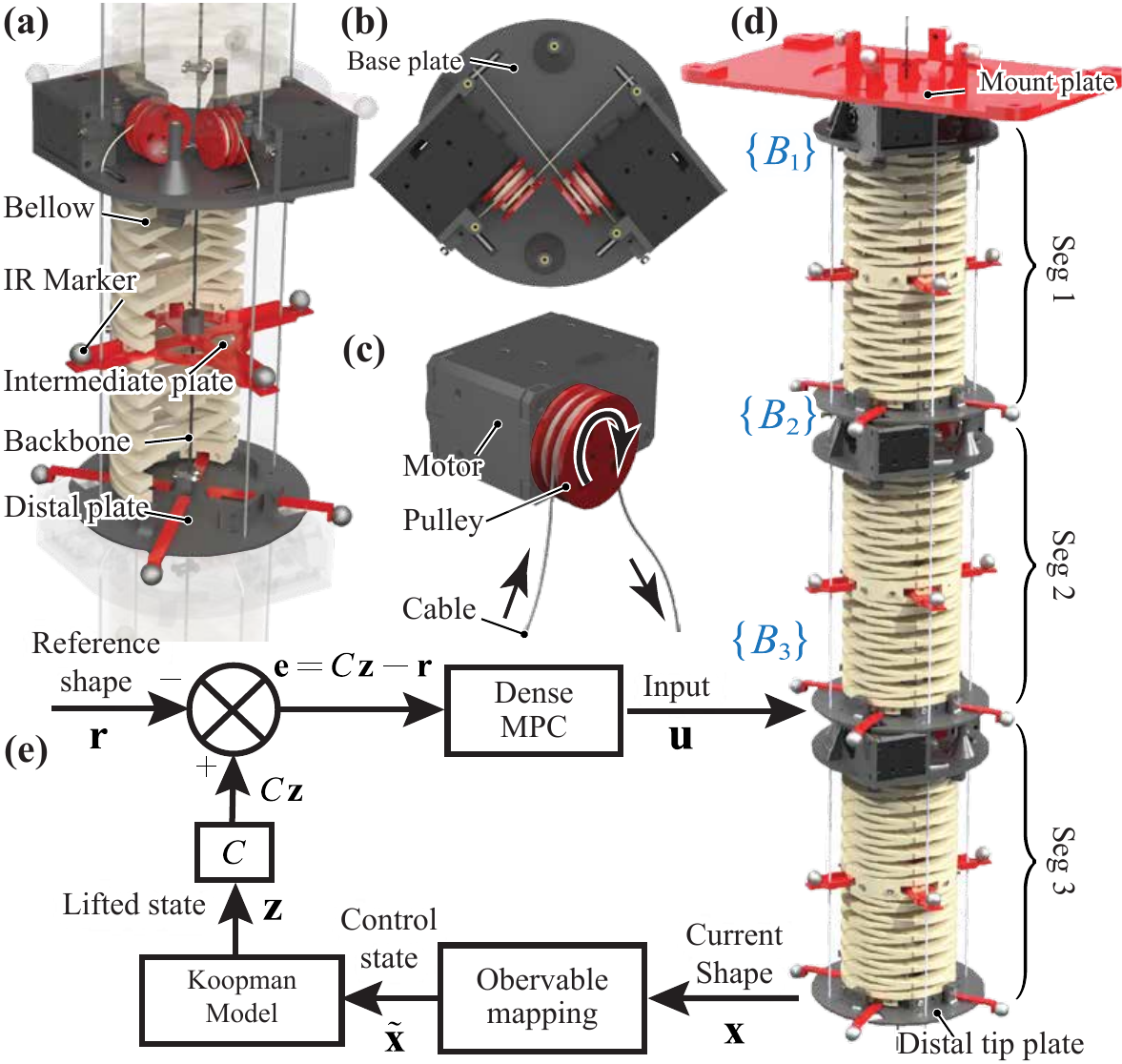}
    \caption{Robot design and experimental setup. 
    (a) Modular design for the robot segment. 
    (b) Base plate with orthogonally mounted motors.
    (c) Motor–pulley–cable transmission with dual-groove winding. 
    (d) Segment indexing and local frame definition. 
    (e) System integration and control diagram.}
    \label{fig:PhysicalSetup}
\end{figure}

\subsection{Robot Design and Data Collection}
\label{subsec: physical_setup}

\begin{table}[t!]
\centering
\caption{Summary of static repeatability and dynamic uncertainty for 3- and 5-segment robots.}
\begin{tabular}{l l c c}
\toprule
\textbf{Robot} & \textbf{Metric} & \textbf{Static (mm$^2$)} & \textbf{Dynamic (mm$^2$)} \\
\midrule

\multirow{2}{*}{3-segment}
& Tip error & $5.60 \pm 4.10$ & $118.53 \pm 122.06$ \\
& Global shape error & $2.35 \pm 3.23$ & $40.21 \pm 40.76$ \\

\midrule

\multirow{2}{*}{5-segment}
& Tip error & $8.47 \pm 30.10$ & $314.65 \pm 312.48$ \\
& Global shape error & $2.49 \pm 6.50$ & $77.76 \pm 261.86$ \\

\bottomrule
\end{tabular}

\vspace{1pt}

\footnotesize
\raggedright
Note: Root-mean-square (RMS) tip errors are approximately 2.37 mm (0.40\%$L$) and 10.89 mm (1.82\%$L$) for the 3-segment robot ($L$=0.6m), and 2.91 mm (0.29\%$L$) and 17.74 mm (1.77\%$L$) for the 5-segment robot ($L$=1.0m), under static and dynamic conditions, respectively.
\label{tab:repeatability_summary}
\end{table}

\begin{figure*}[h!]
    \centering
    \includegraphics[width=1.0\linewidth]{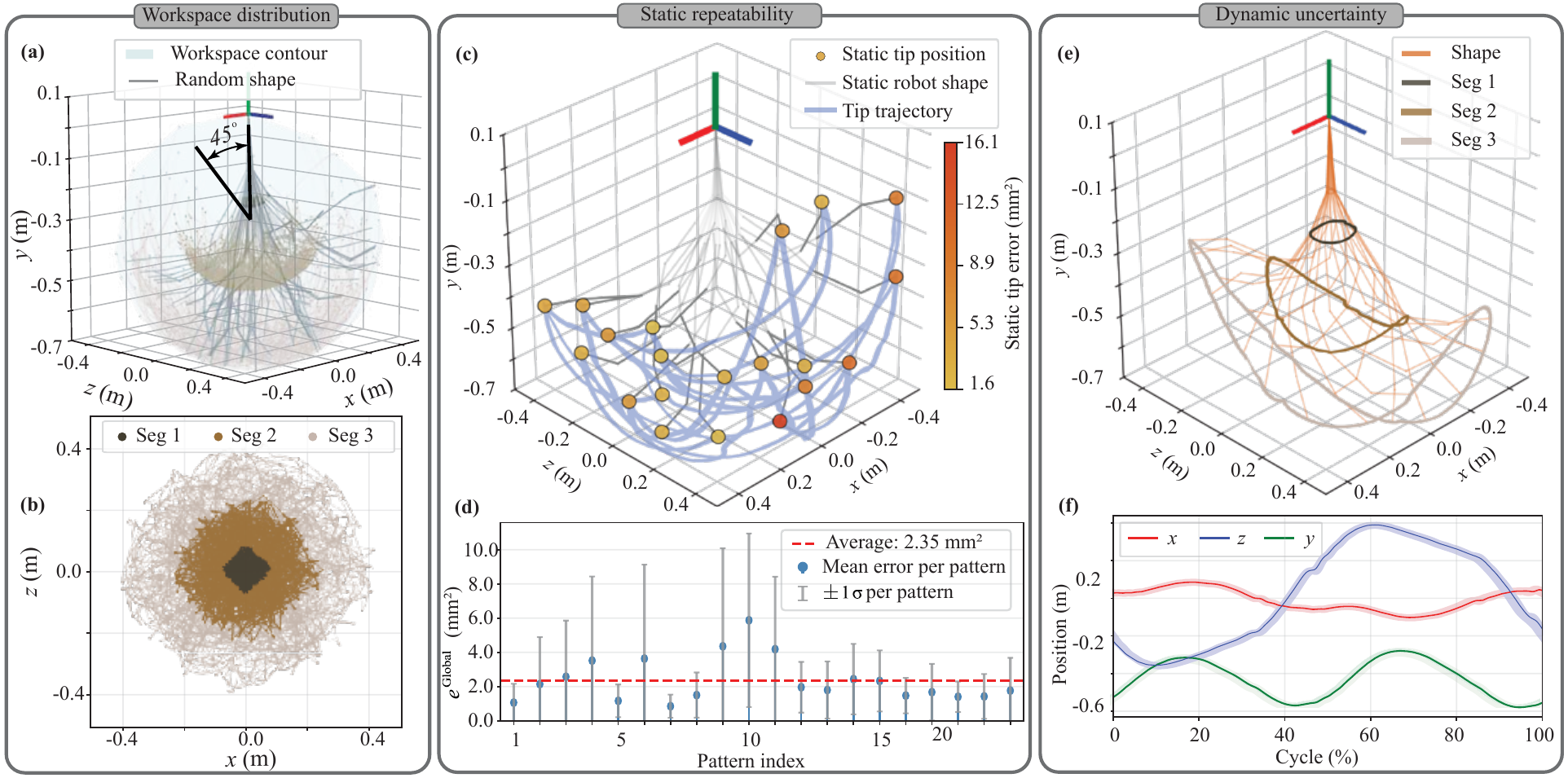}
\caption{Characterization of the 3-segment soft robotic arm. 
(a, b) Workspace and collected data distribution. (a) Three-dimensional reachable workspace under randomized excitation, showing a maximum bending angle of approximately 145$^\circ$. (b) Projection onto the $x$--$z$ plane, illustrating segment-dependent workspace regions. 
(c, d) Static repeatability across 20 sampled configurations. (c) Task-space visualization of robot shapes (gray), final static tip positions (colored markers), and tip trajectories over repeated trials (blue). (d) Global shape error distribution across patterns, showing mean error per pattern, overall average, and $\pm1\sigma$ variation. 
(e, f) Dynamic uncertainty under 20 repeated trajectory execution. (e) Robot shapes sampled at 1~s intervals during motion. (f) Tip position trajectories in $x$, $y$, and $z$ directions over multiple cycles, with shaded regions indicating $\pm1\sigma$ variation. 
 }
    \label{fig:3_seg_analysis}
\end{figure*} 


Unlike conventional cable-driven soft robotic arms that place all actuators at a centralized base \cite{yuan_design_2025,dewi_lightweight_2024}, the proposed system mounts motors on the base of each segment, as shown in Fig.~\ref{fig:PhysicalSetup}~(a). 
Each segment weighs approximately 0.5~kg with a length of $L_{seg}$=0.2~m. 

\begin{figure*}[h!]
    \centering
    \includegraphics[width=1.0\linewidth]{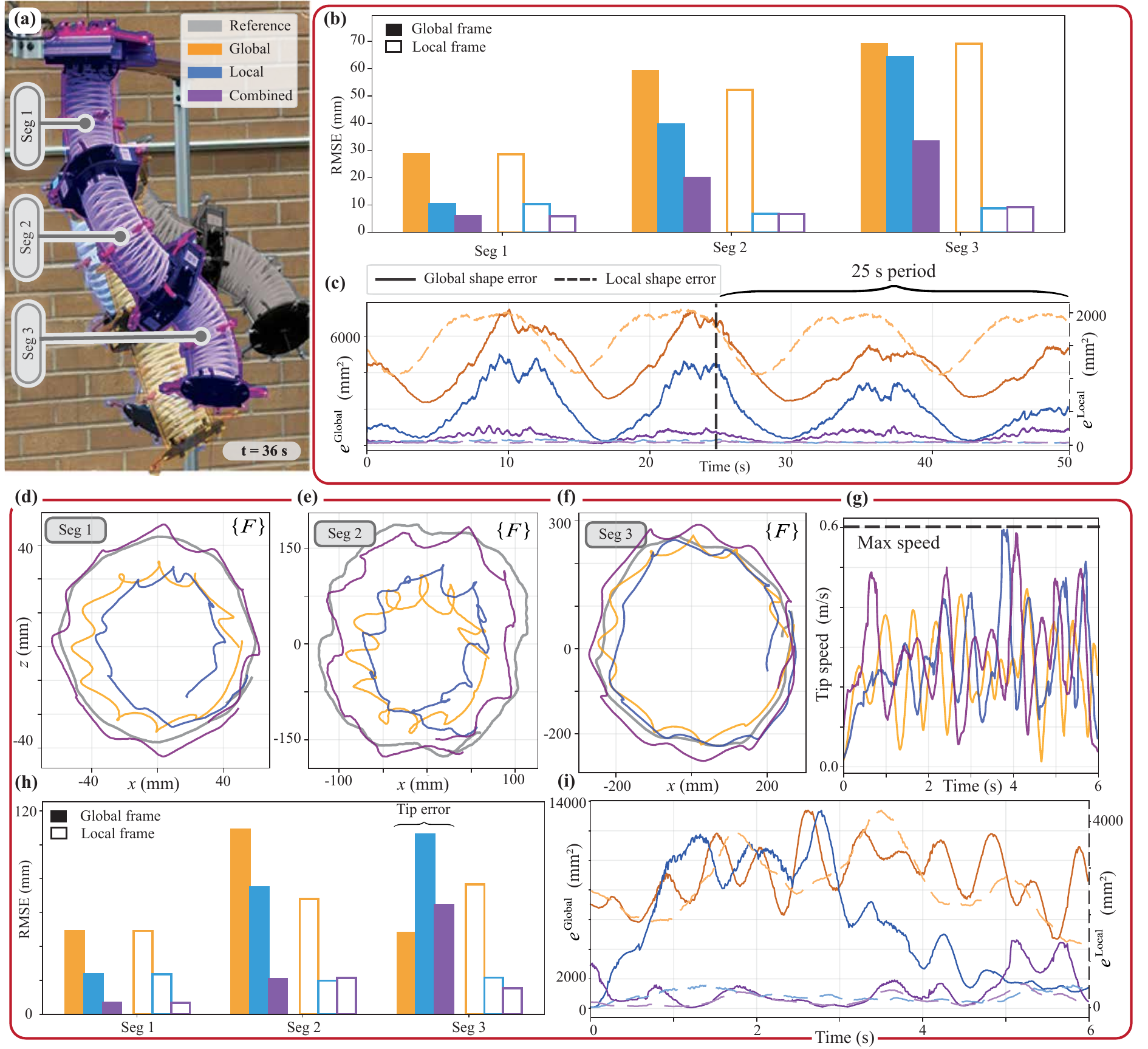}
    \caption{Slow and fast shape tracking comparison for the 3-segment soft robotic arm using global (orange), local (blue), and combined (purple) observables. The reference shape is shown in gray. 
    (a-c) Tracking results for the 25~s-period trajectory (slow shape control):  (a) Physical experiment snapshot at t=36 s.  (b) Segment-wise endpoint tracking RMSE evaluated in both the global and local frames. (c) Time histories of global and local shape error in two periods. 
    (d-i) Tracking results for the 6~s-period reference trajectory (fast shape control):  (d-f) Projected endpoint trajectories of each segment. (g) Tip speed magnitude, with the maximum speed approaching 0.6~m/s. (h) Segment-wise endpoint tracking RMSE. (i) Global and local shape errors over the 6~s-period trajectory. 
    }
    \label{fig:Compare_3seg}
\end{figure*}

\subsubsection{Mechanical Structure} A super-elastic NiTi backbone (diameter $1.8~\mathrm{mm}$) runs continuously through all segments, providing axial support and maintaining alignment between adjacent modules. Surrounding the backbone, each segment consists of two compliant lattice bellows inspired by \cite{yuan_design_2025, guan_trimmed_2023}. The bellows are 3D printed from TPU 64D to enable large bending with smaller torsion and introduce damping. 

\subsubsection{Actuation} Each segment is driven by two motors (Dynamixel XL430-W250-T) mounted orthogonally on the base plate (diameter 120~mm), as shown in Fig.~\ref{fig:PhysicalSetup}~(b). Each motor drives a pair of antagonistic cables (PowerPro 100 lb microfilament braided line). The cables wrap around a dual-groove pulley with a winding diameter of 10~mm (Fig.~\ref{fig:PhysicalSetup}~(c)). The two cables are wound in opposite directions on separate pulley grooves: motor rotation pulls one cable while releasing the other. 
The control input $\Delta \mathbf{u} \in \mathbb{R}^{6}$ is the motor position increments.

\subsubsection{Shape Measurement} The robot's shape is measured at 100~Hz by a motion capture system (OptiTrack Flex 3). For each segment, four infrared (IR) markers define a rigid body on both the intermediate and distal plates (Fig.~\ref{fig:PhysicalSetup}~(a)), which reconstruct discrete backbone poses in the global frame $\{F\}$. An additional marker set is placed on the mount plate to define the local base frame $\{B_1\}$. For the 3-segment robot, 7 rigid-body poses are reconstructed along the backbone (Fig.~\ref{fig:PhysicalSetup}~(d)), yielding the full state $\mathbf{x} \in \mathbb{R}^{49}$. This state is mapped to global and local observables, $\tilde{\mathbf{x}}^{\mathrm{global}}, \tilde{\mathbf{x}}^{\text{local}} \in \mathbb{R}^{6}$, consistent with the observable mapping in Sec.~\ref{subsec:obs_map_design}. 

\subsubsection{System Integration}

The overall system architecture and communication pipeline are illustrated in Fig.~\ref{fig:PhysicalSetup}~(e). Each motor operates with an internal PID controller to track target encoder positions, sent from a host computer via serial communication. ROS2 is used to integrate state reconstruction, reference generation, data synchronization, logging, and real-time K-MPC computation.

\subsubsection{Data Collection}
For data collection, 600,000 data snapshots (1.6 hours) are collected, as shown in Fig.~\ref{fig:3_seg_analysis}~(a, b). The same ramp-and-hold scheme described in Fig.~\ref{fig:SimulationSetup}~(c) is applied to the motor-position inputs. Each input consists of a 1–5~s ramp phase followed by a 1~s hold period. Motor position commands are sampled within predefined bounds that respect motor torque limits and prevent self-contact between segments. 

\subsection{Static Repeatability and Dynamic Uncertainty}
\label{subsec:static_dyn_uncertainty}

To benchmark performance variation, we characterize the intrinsic static repeatability and dynamic uncertainty of the physical system. The quantitative results are summarized in Table~\ref{tab:repeatability_summary}. The evaluation is quantified by tip position error and global shape error metrics introduced in Sec.~\ref{subsec:error_metric}. 

 Static performance of the 3-segment robot is shown in Fig.~\ref{fig:3_seg_analysis}~(c, d). Twenty target inputs are selected using Latin hypercube sampling to ensure uniform coverage of the input space. Each target is executed using a ramp-and-hold scheme (5~s ramp, 5~s hold), and the final measurement during each hold period is used to determine the static equilibrium shape. The sequence is repeated 20 times to evaluate repeatability. Fig.~\ref{fig:3_seg_analysis}~(c) visualizes the robot shapes, repeated tip trajectories, and the corresponding static tip errors. The quantitative repeatability results are summarized in Fig.~\ref{fig:3_seg_analysis}~(d), where the mean global shape error for each sampled configuration, together with the $\pm1\sigma$ variation across repeated trials, is reported. The average global shape error over all 20 configurations is 2.35~mm$^2$.

Dynamic performance is evaluated by executing a 15~s trajectory over 20 repeated cycles, as shown in Fig.~\ref{fig:3_seg_analysis}~(e, f). Fig.~\ref{fig:3_seg_analysis}~(e) illustrates the variation in robot motion across repeated executions, while Fig.~\ref{fig:3_seg_analysis}~(f) shows the mean tip trajectories and their variations in $x$, $y$, and $z$. 

\begin{figure*}[t!]
    \centering
    \includegraphics[width=1\linewidth]{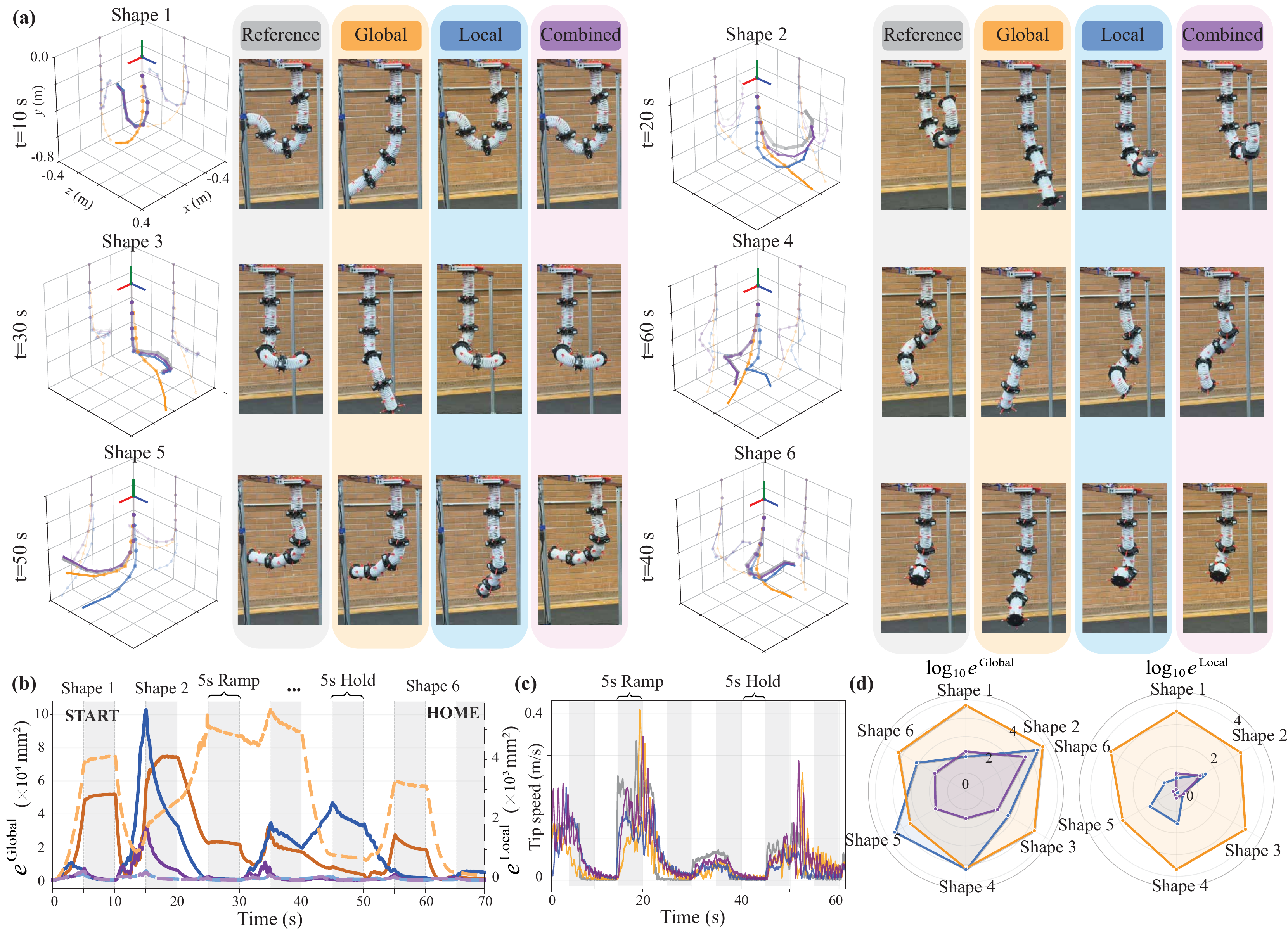}
\caption{Shape tracking performance of the 5-segment soft robotic arm. 
(a) Reference shapes and corresponding tracking results. For each shape, the left subpanel shows the reference and reconstructed backbone, followed by snapshots under global, local, and combined observables.  
(b) Global and local shape errors over time. The combined observable consistently shows lower errors than global and local observables. 
(c) Tip speed magnitude, with the maximum speed approaching 0.4~m/s.
(d) Global and local shape errors for each static shape, shown in $\log_{10}$ scale.
}
    \label{fig:5_seg_demo}
\end{figure*} 

\subsection{Control the 3-Segment Robot}
\label{subsec:exp_3seg_control}

In this subsection, three observables introduced in Sec.~\ref{subsec:obs_map_design} are evaluated for shape control on the 3-segment soft robotic arm. A time-delay coordinate of $N_d=20$ is selected based on multi-step prediction validation \cite{haggerty_control_2023}. The dense MPC controller (Eq.~\eqref{eq:mpc_objective}) operates at 100~Hz with a prediction horizon of $N_H=5$.  All Koopman models are trained on the same dataset and track the same reference trajectory. The shape control performance is quantified by the global and local shape errors introduced in Eq. \eqref{eq:global_error} and Eq. \eqref{eq:local_error}. The segment-wise tracking performance is quantified by  $e^{\{\lambda\}}=\sqrt{\frac{1}{N}\sum_{k=1}^{N}\left\|\mathbf{p}_{k}^{\{\lambda\}}-\mathbf{p}_{k}^{*,\{\lambda\}}\right\|_2^2}$.

We use two different tracking speeds to show that the combined observable provides the best tracking performance under both slow and fast motions. The results highlight the limitations of global and local observables and demonstrate the advantage of the combined observable for shape control.


\subsubsection{Slow shape control}
The tracking results of a slow time-varying shape trajectory (25~s period) are shown in Fig.~\ref{fig:Compare_3seg}~(a)--(c). A representative snapshot is captured in Fig.~\ref{fig:Compare_3seg}~(a). The segment-wise endpoint errors are reported in Fig.~\ref{fig:Compare_3seg}~(b). The global observable (orange) produces the largest errors in the global frame, especially for distal points, and also yields the largest local shape errors. This indicates that the global observable alone does not accurately capture segment-wise deformation. The local observable (blue) reduces the local shape error, but still leaves larger errors in the global frame. In contrast, the combined observable (purple) has the smallest global and local shape error for most of the time, as shown in Fig.~\ref{fig:Compare_3seg}~(c).


\subsubsection{Fast shape control}

To further evaluate performance under faster motion, we track the same shape trajectory with a 6~s period. The projected endpoint trajectories represented in the global frame are shown in Fig.~\ref{fig:Compare_3seg}~(d)--(f). Compared with slow tracking, faster motion results in larger tracking errors for all controllers, and the trajectory exhibits higher-frequency vibration. The corresponding tip speed is shown in Fig.~\ref{fig:Compare_3seg}~(g). The maximum tip speed is approximately 0.6 m/s (i.e., the robot travels its full length in 1 second), where the motors are operating near their speed limitation. The quantitative tracking performance is summarized in Fig.~\ref{fig:Compare_3seg}~(h). Although the robot speed is increased, the quantitative results in Fig.~\ref{fig:Compare_3seg}~(h) show the same trend in Seg~1 and Seg~2: the combined observable provides lower errors than the global or local observable alone. For Seg~3, the global observable achieves the lowest RMSE for the tip error in the global frame. However, its global and local shape errors remain large, as shown in Fig.~\ref{fig:Compare_3seg}~(i). This confirms that tip error alone cannot determine the multi-segment robot shape, since a controller may reduce the tip error while still producing incorrect deformation along the backbone (as introduced in Fig.~\ref{fig:error_compare}). 

\subsection{Control the 5-Segment Robot}
\label{subsec:exp_5seg_control}

Compared to the 3-segment system, the 5-segment soft robotic arm exhibits stronger nonlinearity, higher dimensionality, and increased uncertainties (introduced in Sec.~\ref{subsec:static_dyn_uncertainty}). By evaluating shape control across multiple reference shapes, we demonstrate that the proposed combined observable extends beyond simple deformation patterns and maintains control accuracy as the segment number increases. 

Six distinct target shapes are intentionally chosen to span diverse deformation patterns across the robot workspace, as shown in Fig.~\ref{fig:5_seg_demo}~(a). These shapes include multi-curvature bending, asymmetric deformation, and helix. We execute all six shapes in a single trail. The reference trajectory is generated using a sequential ramp-and-hold scheme: the robot transitions to each shape over a 5~s ramp phase and then holds  for 5~s before moving to the next shape, as shown in
Fig.~\ref{fig:5_seg_demo}~(b). The white and gray regions indicate the ramp and hold phases.

For large and asymmetric deformations (Shapes~1, 3, and 6), local observable achieves significantly lower local shape errors
, while maintaining moderate global shape errors. 
In contrast, global observable fails to capture internal deformation, resulting in much larger global shape errors 
and local shape errors exceeding 718.63~mm$^2$. 

For Shapes~2 and~5, which are dominated by simple large bending patterns, the global observable achieves relatively lower global shape error than the local observable. 
This suggests that simple bending can be approximated more effectively by the global observable, whereas the local observable suffers from accumulated error due to the lack of explicit global position information. However, global observable still underfits internal deformation, yielding larger local shape errors 
than local observable. 

The corresponding tip speed is shown in Fig.~\ref{fig:5_seg_demo}~(c). The maximum tip speed is approximately 0.4 m/s. It should be noted that the oscillatory velocity profile captures dynamic effects arising from the robot's inertia (total weight 2.5 kg) and shape deformation. During the holding phases, the tip speed decays toward zero, confirming that the robot reaches an equilibrium state.

Across all six shapes, the combined observable provides the best tracking performance, maintaining both global and local shape errors within relatively low ranges, with an average global shape error of 2739.92~mm$^2$ (5.23\%$L$), and local shape error of 52.19~mm$^2$ (3.61\%$L_{seg}$). Notably, the combined observable is structurally consistent rather than shape-specific. The static global and local shape errors for all six shapes are summarized in Fig.~\ref{fig:5_seg_demo}~(d).

\begin{figure}[t!]
    \centering
    \includegraphics[width=1.0\linewidth]{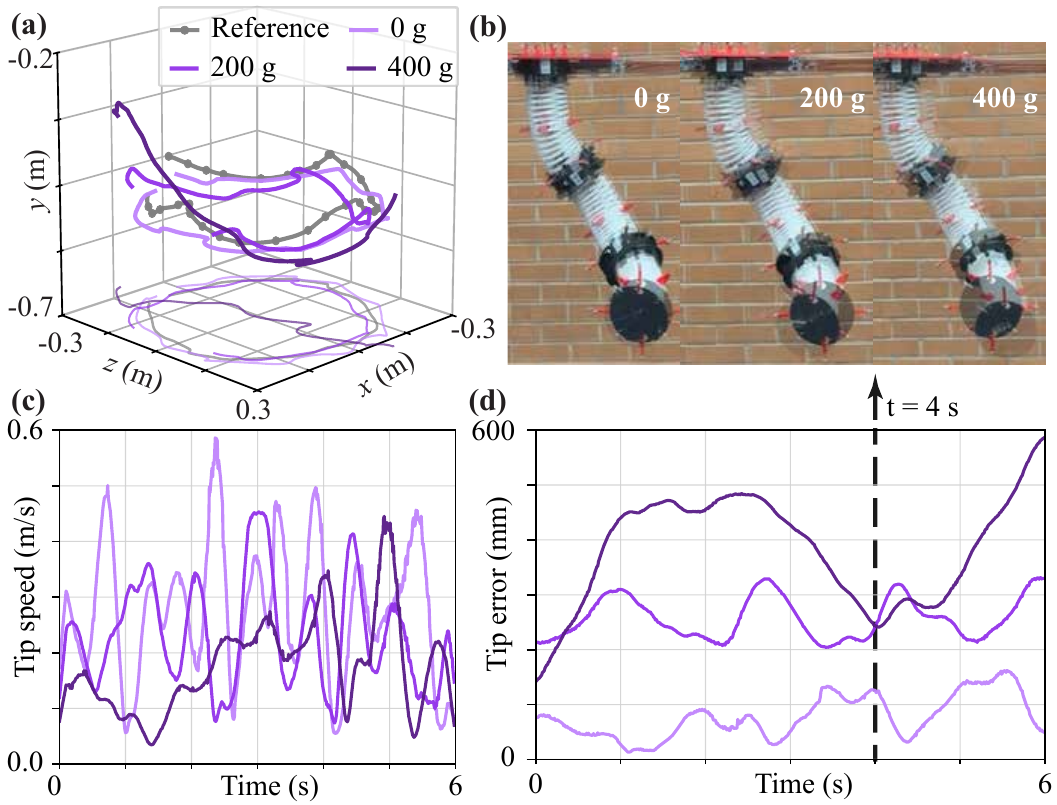}
    \caption{Distal payload experiment on the 3-segment soft robotic arm under three payload conditions: 0~g, 200~g, and 400~g. (a) Tip trajectories and their planar projections. (b) Representative experimental snapshots at $t=4$~s for the three payload conditions. (c) Tip speed magnitude over time. (d) Tip error over time.}
    \label{fig:Tipload_3seg}
\end{figure}

\begin{figure}[t!]
    \centering
    \includegraphics[width=1.0\linewidth]{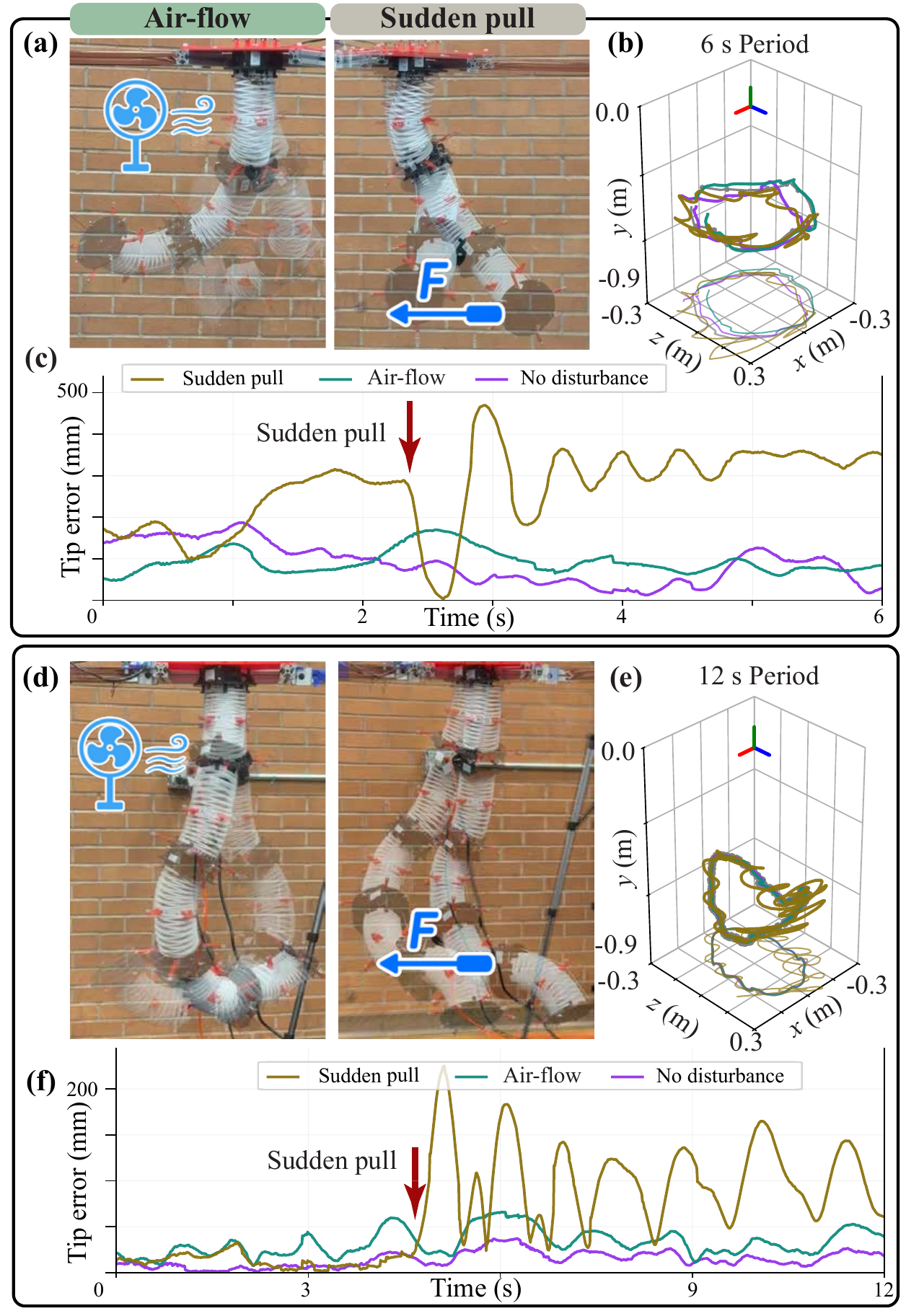}
\caption{Robustness evaluation of the proposed combined observable controller under unknown external disturbances. (a-c) Results for the 3-segment soft robotic arm tracking the 6~s-period trajectory. (a) Representative snapshots of the air-flow and sudden-pull disturbances. (b) Tip trajectories under different disturbance conditions and their projection. (c) Tip error over time. (d-f) Results for the 5-segment soft robotic arm tracking the 12~s-period trajectory. (d) Representative disturbance snapshots. (e) Tip trajectories under different disturbance conditions and their projection. (f) Tip error over time. 
}
    \label{fig:suddenpull_windblow}
\end{figure}

\subsection{Robustness Experiment}
\label{subsec:robustness}

To further evaluate the proposed controller remaining robust beyond free-deformation, we test the combined observable controller under three types of unmodeled external effects: distal payloads, air-flow, and sudden-pull without retraining. The results show that although additional disturbances increase the tracking error, the combined observable controller remains stable and continues to execute the time-varying shape control.

\subsubsection{Distal Payload} 
\label{subsubsec:distal_payload} To evaluate robustness under unmodeled payloads, we attach additional weights to the tip of the 3-segment robot and repeat the fast tracking task. Two payload conditions are tested: 200 g (13.1\% of robot weight) and 400 g (26.1\% of robot weight). The Koopman model is trained using the original no-payload dataset, and no additional payload-specific training data are used. 
Fig.~\ref{fig:Tipload_3seg}~(a) compares the tip trajectories under different payload conditions. As the payload increases, the measured trajectory deviates farther from the reference, indicating that the additional distal load introduces significant unmodeled deformation. Representative snapshots at $t=4$~s are shown in Fig.~\ref{fig:Tipload_3seg}~(b), where the heavier payload produces a larger downward deflection.
The corresponding tip speed and tip tracking error are shown in Fig.~\ref{fig:Tipload_3seg}~(c) and (d), respectively. Under the no-payload condition, the robot tracks the dynamic reference trajectory with a mean tip error of 86.9~mm and a mean tip velocity of 0.28~m/s. With a 200~g payload, the mean tip error increases to 259.9~mm, while the mean tip velocity decreases to 0.23~m/s. With a 400~g payload, the mean tip error further increases to 394.5~mm, and the mean tip velocity decreases to 0.17~m/s. These results show that increasing the distal payload increases the tip tracking error and reduces the achievable motion speed. Nevertheless, the controller remains stable and continues to execute the time-varying shape-tracking task without payload-specific retraining.

\begin{figure*}[t!]
    \centering
    \includegraphics[width=1.0\linewidth]{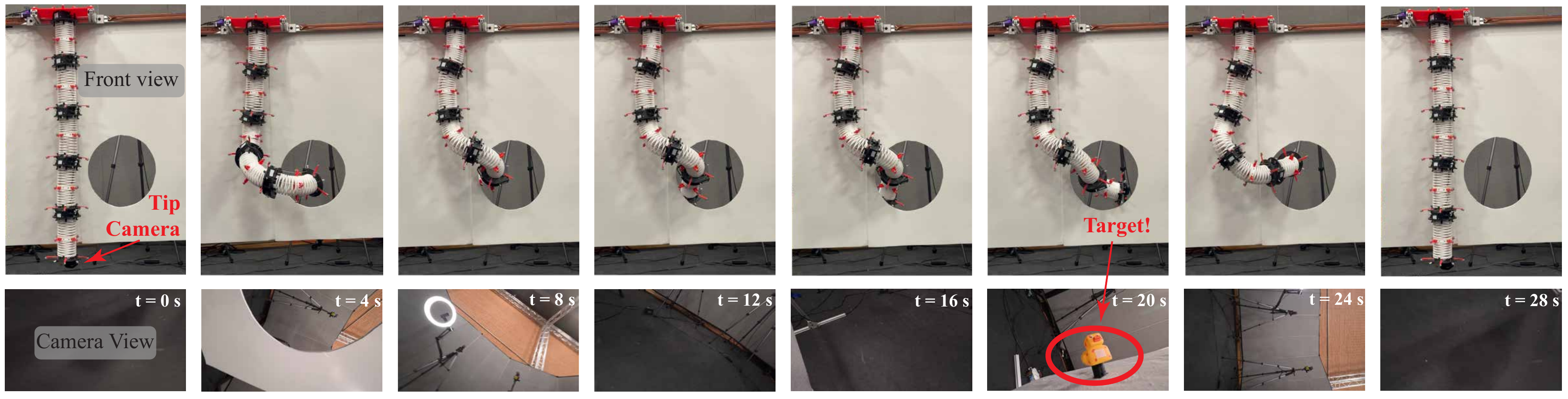}
\caption{Confined-space shape-control demonstration using the 5-segment soft robotic arm with the proposed combined observable controller.}
    \label{fig:demo_5seg}
\end{figure*}

\subsubsection{External Disturbances}

To evaluate robustness against unknown external disturbances, the proposed combined observable controller is tested under two conditions: continuous air-flow and sudden lateral pulling. The airflow is generated by a fan placed at a fixed distance from the robot. The sudden-pull disturbance is applied near the distal body through a magnetic attachment, producing a peak lateral force of approximately 7~N (46.7\% of the robot's weight).
Figs.~\ref{fig:suddenpull_windblow}~(a) and (d) show representative snapshots of the two disturbance scenarios for the 3- and 5-segment robots, respectively. Figs.~\ref{fig:suddenpull_windblow}~(b) and (e) compare the corresponding tip trajectories under the no-disturbance, air-flow, and sudden-pull conditions. For both robots, the air-flow disturbance causes only a slight deviation, whereas the sudden pull produces a noticeable transient deviation.
The corresponding tip tracking errors are shown in Figs.~\ref{fig:suddenpull_windblow}~(c) and (f). For the 3-segment robot tracking the 6~s-period trajectory, the mean tip error increases only slightly from 89.4~mm to 93.1~mm under continuous air flow. In contrast, the sudden pull increases the mean tip error to 297.1~mm, producing a sharp transient response before the robot converges back toward the desired trajectory.
For the 5-segment robot tracking the 12~s-period trajectory, the mean tip error increases from 14.3~mm under nominal conditions to 32.6~mm under continuous air flow and to 74.8~mm under the sudden-pull disturbance. Compared with the 3-segment robot, the longer structure exhibits a larger transient response due to its increased inertia and distributed flexibility, although the controller remains stable and gradually recovers after the disturbance is removed.

Overall, the proposed controller maintains stable shape control under both continuous and impulsive external disturbances. Although unknown disturbances increase the tracking error, the controller consistently recovers the desired motion without retraining or disturbance-specific modeling.

\subsection{Confined-Space Shape-Control Demonstration}
\label{subsec:confined_demo}

To illustrate the applicability of the proposed shape controller beyond free-space tracking, we demonstrate a confined-space inspection task using the 5-segment soft robotic arm. 

In this experiment, the robot follows a pre-generated sequence of reference shapes that guides the body through a circular aperture toward a target located on the opposite side of the constraint. As shown in Fig.~\ref{fig:demo_5seg}, the robot starts from a straight configuration and gradually bends its distal segments toward the aperture. The front-view images illustrate the global body deformation, while the camera-view images provide the local view during execution. At t=20~s, the target becomes visible in the tip-camera view, indicating that the planned body deformation has been successfully executed while maintaining the desired overall shape. The complete demonstration lasts 28~s, covering the approach, aperture traversal, target observation, and return motion.

Although the reference shape sequence is manually specified, the proposed controller can naturally serve as the low-level tracking module for planning frameworks \cite{yang_combined_2024,yuan_hybrid_2026,wu_novel_2022, shentu_sampling-based_2026}. Therefore, the controller is compatible with future shape-planning and obstacle-avoidance algorithms.

\section{Discussion}
\label{sec:discussion}

The results presented in this work highlight that the primary challenge in multi-segment soft robot shape control is not solely model accuracy, but also the choice of control state and cost function. This suggests that, for multi-segment soft robotic systems, the design of observable mappings plays a more critical role than increasing model complexity. The following subsections analyze the underlying mechanisms, provide supporting evidence, and discuss the implications and limitations of the proposed approach.

\subsection{Choice of local observable} \label{subsubsec:lifting_functions_explanation}
Most existing Koopman-based control approaches for soft robotic arms rely on \textit{global observables}, i.e., nonlinear functions of position coordinates measured in the global frame \cite{bruder_modeling_2019,bruder_koopman-based_2021,bruder_data-driven_2021,bruder_koopman-based_2024,haggerty_control_2023,singh_controlling_2023}. Our results in Sec.~\ref{sec:numerical} show that it is insufficient for shape control in multi-segment systems. In particular, inaccuracies in modeling the extension direction lead to poor representation of local deformation and degraded control performance. Despite this limitation, global observables remain effective under small deflections or in single-segment systems \cite{bruder_data-driven_2021, haggerty_control_2023}. In such cases, the backbone deformation is well approximated by constant-curvature models \cite{della_santina_improved_2020}, and the end-effector position can be reasonably represented by its projection onto the plane perpendicular to the extension direction. Under this assumption, the mapping from the projected coordinates to the end-effector position is approximately unique.

\begin{figure}[t!]
    \centering
    \includegraphics[width=0.9 \linewidth]{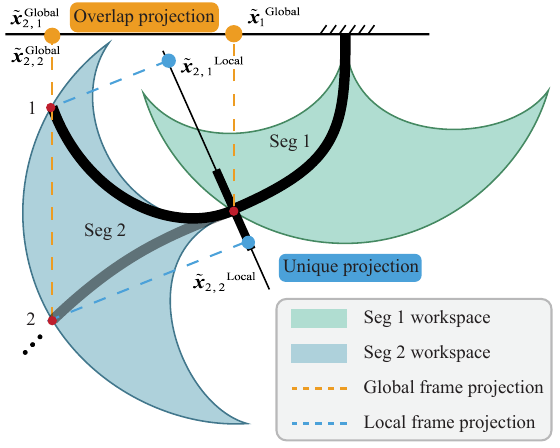}
    \caption{Illustration of global and local observables. The workspace is shown in green (Seg~1) and blue (Seg~2). Orange dashed lines denote projections onto the global normal plane, where multiple configurations of Seg~2 map to the same projected coordinates (overlap projection). In contrast, blue dashed lines denote projections in local segment frames, where each configuration is uniquely represented (unique projection).}
    \label{fig:local_frame_explainer}
\end{figure}

However, this assumption breaks down in multi-segment systems. As illustrated in Fig.~\ref{fig:local_frame_explainer}, global frame projections exhibit \textit{overlap projection}: multiple backbone configurations can produce the same projected coordinates. This overlap leads to a non-unique mapping between the observable and the robot shape, introducing ambiguity in the state representation and resulting in control inaccuracies, as observed in Sec.~\ref{subsec:exp_3seg_control}.

To address this ambiguity, we project observables in local segment frames. Then the mapping between coordinates and segment deformation becomes locally unique. This \textit{unique projection} preserves segment-wise geometry and avoids the overlap observed in global coordinates. This motivates the use of \textit{local observables}, which better capture segment-wise geometry and improve modeling accuracy for multi-segment soft robotic arms.

\subsection{Shape Error Metric}

The choice of shape error metric (introduced in Sec. \ref{subsec:error_metric}) is critical for evaluating the tracking performance of multi-segment soft robotic arms and for designing the controller objective function. Existing methods typically emphasize global shape error (task-space alignment) due to sensing or modeling limitations.

Our results show that, in real implementation, optimizing either metric alone is insufficient. Minimizing global shape error does not guarantee accurate local deformation, while minimizing local shape error can lead to significant drift in task space (as introduced in Fig.~\ref{fig:error_compare}). For example, in Shape~4 (Fig.~\ref{fig:5_seg_demo}), global observables achieve lower global shape error than local observables, yet fail to capture the local deformation. Conversely, local observables produce accurate deformations but accumulate global shape error along the arc-length. These results indicate that effective multi-segment shape control requires consideration of both global and local shape errors. 

\subsection{Limitations}

Despite these encouraging results, several limitations remain.  
First, the current implementation relies on external motion-capture measurements, which constrain the system to laboratory environments. 
Second, reference trajectories are generated from real data without planning, thereby restricting the controller to a predefined shape rather than a fully autonomous trajectory. 
Third, the training data must be carefully curated to avoid self-contact, limiting the explored state space. 
Fourth, the achievable tracking performance is constrained by the motion-capture update rate and actuator capability. More aggressive excitation leads to increased oscillation and reduced convergence stability.
As a result, dynamic shape control for multi-segment soft robotic arms in the highly-inertial regime remains an open challenge for the current hardware.

Additionally, the experimental validation focuses on free deformation shape control and robustness tests under moderate disturbances. As a fixed-model control framework, the proposed method may degrade when the robot dynamics change significantly, such as under heavy payloads or strong external forces. Future work could combine the proposed method with adaptive Koopman control~\cite{bruder_koopman-based_2021} to improve robustness across different loading conditions.

\section{Conclusion}
\label{sec:conclusion}

In this paper, we propose a Koopman-based dense-MPC framework for real-time shape control of multi-segment soft robotic arms and validate it through numerical and physical experiments. By combining global and local observables within a unified Koopman control architecture, the proposed approach reduces both global and local shape errors during shape control. Although further scaling to higher-dimensional physical systems is limited by motor capacity and space constraints, the simulation results demonstrate the scalability of the method with up to 10 independently actuated segments, indicating strong potential to extend the combined observable approach to more complex multi-segment shape control.

Future work will focus on incorporating richer sensing modalities, such as strain-based parameterizations and fiber Bragg grating (FBG) sensors, to enable higher-frequency and fully onboard control. Extending the framework to account for self-contact, external payloads, and force-based actuation will further improve robustness and applicability in complex manipulation tasks.

\bibliographystyle{IEEEtranN}
\bibliography{bst,refs}

@IEEEtranBSTCTL{IEEEexample:BSTcontrol,
    CTLuse_url = "no",
}

@article{bruder_koopman-based_2021,
	title = {Koopman-{Based} {Control} of a {Soft} {Continuum} {Manipulator} {Under} {Variable} {Loading} {Conditions}},
	volume = {6},
	copyright = {https://ieeexplore.ieee.org/Xplorehelp/downloads/license-information/IEEE.html},
	issn = {2377-3766, 2377-3774},
	url = {https://ieeexplore.ieee.org/document/9477047/},
	doi = {10.1109/LRA.2021.3095268},
	language = {en},
	number = {4},
	urldate = {2025-01-23},
	journal = {IEEE Robotics and Automation Letters},
	author = {Bruder, Daniel and Fu, Xun and Gillespie, R. Brent and Remy, C. David and Vasudevan, Ram},
	month = oct,
	year = {2021},
	pages = {6852--6859},
}

@article{bruder_koopman-based_2024,
	title = {A {Koopman}-based residual modeling approach for the control of a soft robot arm},
	issn = {0278-3649, 1741-3176},
	url = {https://journals.sagepub.com/doi/10.1177/02783649241272114},
	doi = {10.1177/02783649241272114},
	language = {en},
	urldate = {2025-01-05},
	journal = {The International Journal of Robotics Research},
	author = {Bruder, Daniel and Bombara, David and Wood, Robert J},
	month = oct,
	year = {2024},
	pages = {02783649241272114},
}

@article{li_position_2025,
	title = {Position and {Orientation} {Tracking} {Control} of a {Cable}-{Driven} {Tensegrity} {Continuum} {Robot}},
	volume = {41},
	copyright = {https://ieeexplore.ieee.org/Xplorehelp/downloads/license-information/IEEE.html},
	issn = {1552-3098, 1941-0468},
	url = {https://ieeexplore.ieee.org/document/10891749/},
	doi = {10.1109/TRO.2025.3543292},
	urldate = {2025-05-07},
	journal = {IEEE Transactions on Robotics},
	author = {Li, Fei and Yang, Hao and Gu, Guoying and Wang, Yongqing and Peng, Haijun},
	year = {2025},
	pages = {1791--1811},
}

@article{yuan_design_2025,
	title = {Design of a {Deployable} {Continuum} {Robot} ({DCR}) {With} {Coiling} {Mechanism}},
	copyright = {https://ieeexplore.ieee.org/Xplorehelp/downloads/license-information/IEEE.html},
	issn = {1083-4435, 1941-014X},
	url = {https://ieeexplore.ieee.org/document/11026100/},
	doi = {10.1109/TMECH.2025.3572012},
	urldate = {2025-06-27},
	journal = {IEEE/ASME Transactions on Mechatronics},
	author = {Yuan, Peikang and Sun, Changchao and Chang, Xiang and Dong, Xin and Song, Zhibin and Sun, Tao and Dai, Jian S and Kang, Rongjie},
	year = {2025},
	pages = {1--13},
}

@article{chen_versatile_2025,
	title = {A {Versatile} {Neural} {Network} {Configuration} {Space} {Planning} and {Control} {Strategy} for {Modular} {Soft} {Robot} {Arms}},
	volume = {41},
	copyright = {https://ieeexplore.ieee.org/Xplorehelp/downloads/license-information/IEEE.html},
	issn = {1552-3098, 1941-0468},
	url = {https://ieeexplore.ieee.org/document/11049035/},
	doi = {10.1109/TRO.2025.3582807},
	urldate = {2025-12-17},
	journal = {IEEE Transactions on Robotics},
	author = {Chen, Zixi and Guan, Qinghua and Hughes, Josie and Menciassi, Arianna and Stefanini, Cesare},
	year = {2025},
	pages = {4269--4282},
}

@ARTICLE{zhang_stochastic_2025,
  author={Zhang, Guoqing and Wang, Long},
  journal={IEEE Transactions on Robotics}, 
  title={Stochastic Adaptive Estimation in Polynomial Curvature Shape State Space for Continuum Robots}, 
  year={2026},
  volume={42},
  number={},
  pages={261-280},
  doi={10.1109/TRO.2025.3637147}}

@article{yu_data-efficient_2025,
	title = {Data-{Efficient} and {Predefined}-{Time} {Stable} {Control} for {Continuum} {Robots}},
	copyright = {https://ieeexplore.ieee.org/Xplorehelp/downloads/license-information/IEEE.html},
	issn = {1552-3098, 1941-0468},
	url = {https://ieeexplore.ieee.org/document/11301632/},
	doi = {10.1109/TRO.2025.3644946},
	language = {en},
	urldate = {2025-12-26},
	journal = {IEEE Transactions on Robotics},
	author = {Yu, Peng and Liang, Zhenhan and Tan, Ning},
	year = {2025},
	pages = {1--19},
}

@article{coad_vine_2020,
	title = {Vine {Robots}: {Design}, {Teleoperation}, and {Deployment} for {Navigation} and {Exploration}},
	volume = {27},
	issn = {1070-9932, 1558-223X},
	shorttitle = {Vine {Robots}},
	url = {http://arxiv.org/abs/1903.00069},
	doi = {10.1109/MRA.2019.2947538},
	language = {en},
	number = {3},
	urldate = {2026-01-04},
	journal = {IEEE Robotics \& Automation Magazine},
	author = {Coad, Margaret M. and Blumenschein, Laura H. and Cutler, Sadie and Zepeda, Javier A. Reyna and Naclerio, Nicholas D. and El-Hussieny, Haitham and Mehmood, Usman and Ryu, Jee-Hwan and Hawkes, Elliot W. and Okamura, Allison M.},
	month = sep,
	year = {2020},
	note = {arXiv:1903.00069 [cs]},
	pages = {120--132},
}

@misc{zuo_umarm_2025,
	title = {{UMArm}: {Untethered}, {Modular}, {Portable}, {Soft} {Pneumatic} {Arm}},
	shorttitle = {{UMArm}},
	url = {http://arxiv.org/abs/2505.11476},
	doi = {10.48550/arXiv.2505.11476},
	language = {en},
	urldate = {2026-01-04},
	publisher = {arXiv},
	author = {Zuo, Runze and Han, Dong Heon and Li, Richard and Jamal, Saima and Bruder, Daniel},
	month = dec,
	year = {2025},
	note = {arXiv:2505.11476 [cs]},
}

@article{iqbal_continuum_2025,
	title = {Continuum and {Soft} {Robots} in {Minimally} {Invasive} {Surgery}: {A} {Systematic} {Review}},
	volume = {13},
	copyright = {https://creativecommons.org/licenses/by-nc-nd/4.0/},
	issn = {2169-3536},
	shorttitle = {Continuum and {Soft} {Robots} in {Minimally} {Invasive} {Surgery}},
	url = {https://ieeexplore.ieee.org/document/10856003/},
	doi = {10.1109/ACCESS.2025.3535677},
	language = {en},
	urldate = {2026-01-04},
	journal = {IEEE Access},
	author = {Iqbal, Fahad and Esfandiari, Mojtaba and Amirkhani, Golchehr and Hoshyarmanesh, Hamidreza and Lama, Sanju and Tavakoli, Mahdi and Sutherland, Garnette R.},
	year = {2025},
	pages = {24053--24079},
}

@article{kulkarni_soft_2025,
	title = {Soft robots built for extreme environments},
	volume = {5},
	issn = {2769-5441},
	url = {https://www.oaepublish.com/articles/ss.2023.51},
	doi = {10.20517/ss.2023.51},
	language = {en},
	number = {1},
	urldate = {2026-01-04},
	journal = {Soft Science},
	author = {Kulkarni, Mayura and Edward, Sandra and Golecki, Thomas and Kaehr, Bryan and Golecki, Holly},
	month = feb,
	year = {2025},
}

@article{rus_design_2015,
	title = {Design, fabrication and control of soft robots},
	volume = {521},
	issn = {0028-0836, 1476-4687},
	url = {https://www.nature.com/articles/nature14543},
	doi = {10.1038/nature14543},
	language = {en},
	number = {7553},
	urldate = {2026-01-04},
	journal = {Nature},
	author = {Rus, Daniela and Tolley, Michael T.},
	month = may,
	year = {2015},
	pages = {467--475},
}

@article{almanzor_static_2023,
	title = {Static {Shape} {Control} of {Soft} {Continuum} {Robots} {Using} {Deep} {Visual} {Inverse} {Kinematic} {Models}},
	volume = {39},
	copyright = {https://ieeexplore.ieee.org/Xplorehelp/downloads/license-information/IEEE.html},
	issn = {1552-3098, 1941-0468},
	url = {https://ieeexplore.ieee.org/document/10144108/},
	doi = {10.1109/TRO.2023.3275375},
	number = {4},
	urldate = {2026-01-06},
	journal = {IEEE Transactions on Robotics},
	author = {Almanzor, Elijah and Ye, Fan and Shi, Jialei and Thuruthel, Thomas George and Wurdemann, Helge A. and Iida, Fumiya},
	month = aug,
	year = {2023},
	pages = {2973--2988},
}

@article{della_santina_model-based_2020,
	title = {Model-based dynamic feedback control of a planar soft robot: trajectory tracking and interaction with the environment},
	volume = {39},
	issn = {0278-3649, 1741-3176},
	shorttitle = {Model-based dynamic feedback control of a planar soft robot},
	url = {https://journals.sagepub.com/doi/10.1177/0278364919897292},
	doi = {10.1177/0278364919897292},
	number = {4},
	urldate = {2026-01-06},
	journal = {The International Journal of Robotics Research},
	author = {Della Santina, Cosimo and Katzschmann, Robert K and Bicchi, Antonio and Rus, Daniela},
	month = mar,
	year = {2020},
	pages = {490--513},
}

@article{wang_dynamic_2021,
	title = {Dynamic {Control} of {Multisection} {Three}-{Dimensional} {Continuum} {Manipulators} {Based} on {Virtual} {Discrete}-{Jointed} {Robot} {Models}},
	volume = {26},
	copyright = {https://ieeexplore.ieee.org/Xplorehelp/downloads/license-information/IEEE.html},
	issn = {1083-4435, 1941-014X},
	url = {https://ieeexplore.ieee.org/document/9107505/},
	doi = {10.1109/TMECH.2020.2999847},
	language = {en},
	number = {2},
	urldate = {2026-01-06},
	journal = {IEEE/ASME Transactions on Mechatronics},
	author = {Wang, Chengshi and Frazelle, Chase G. and Wagner, John R. and Walker, Ian D.},
	month = apr,
	year = {2021},
	pages = {777--788},
}

@article{lai_constrained_2022,
	title = {Constrained {Motion} {Planning} of a {Cable}-{Driven} {Soft} {Robot} {With} {Compressible} {Curvature} {Modeling}},
	volume = {7},
	copyright = {https://ieeexplore.ieee.org/Xplorehelp/downloads/license-information/IEEE.html},
	issn = {2377-3766, 2377-3774},
	url = {https://ieeexplore.ieee.org/document/9716747/},
	doi = {10.1109/LRA.2022.3152318},
	number = {2},
	urldate = {2026-01-06},
	journal = {IEEE Robotics and Automation Letters},
	author = {Lai, Jiewen and Lu, Bo and Zhao, Qingxiang and Chu, Henry K.},
	month = apr,
	year = {2022},
	pages = {4813--4820},
}

@inproceedings{wang_rl-based_2025,
	address = {Lausanne, Switzerland},
	title = {{RL}-based {Shape} {Control} of {Rod}-driven {Soft} {Arms} {Using} {Strain} {Models}},
	copyright = {https://doi.org/10.15223/policy-029},
	isbn = {979-8-3315-2020-5},
	url = {https://ieeexplore.ieee.org/document/11020890/},
	doi = {10.1109/RoboSoft63089.2025.11020890},
	urldate = {2026-01-06},
	booktitle = {2025 {IEEE} 8th {International} {Conference} on {Soft} {Robotics} ({RoboSoft})},
	publisher = {IEEE},
	author = {Wang, Peiyi and Zeng, Yaxin and Sun, Yuchen and Ni, Zhenwei and Nazeer, Muhammad Sunny and Laschi, Cecilia},
	month = apr,
	year = {2025},
	pages = {1--7},
}

@misc{adibnazari_dynamic_2025,
	title = {Dynamic {Shape} {Control} of {Soft} {Robots} {Enabled} by {Data}-{Driven} {Model} {Reduction}},
	url = {http://arxiv.org/abs/2511.03931},
	doi = {10.48550/arXiv.2511.03931},
	urldate = {2026-01-06},
	publisher = {arXiv},
	author = {Adibnazari, Iman and Sharma, Harsh and Park, Myungsun and Cervera-Torralba, Jacobo and Kramer, Boris and Tolley, Michael T.},
	month = nov,
	year = {2025},
	note = {arXiv:2511.03931 [cs]},
}

@article{shen_online_2024,
	title = {Online {Learning} {Based} {Shape} {Control} for a {Soft} {Manipulator} {Based} on {Spatial} {Features} {Feedback}},
	volume = {9},
	copyright = {https://ieeexplore.ieee.org/Xplorehelp/downloads/license-information/IEEE.html},
	issn = {2377-3766, 2377-3774},
	url = {https://ieeexplore.ieee.org/document/10697303/},
	doi = {10.1109/LRA.2024.3469824},
	number = {11},
	urldate = {2026-01-06},
	journal = {IEEE Robotics and Automation Letters},
	author = {Shen, Yi and Zhang, Jinghao and Yuan, Ye and Zhang, Fumin and Ding, Han},
	month = nov,
	year = {2024},
	pages = {10081--10088},
}

@inproceedings{pei_imu_2024,
	address = {Abu Dhabi, United Arab Emirates},
	title = {{IMU} {Based} {Pose} {Reconstruction} and {Closed}-loop {Control} for {Soft} {Robotic} {Arms}},
	copyright = {https://doi.org/10.15223/policy-029},
	isbn = {979-8-3503-7770-5},
	url = {https://ieeexplore.ieee.org/document/10802377/},
	doi = {10.1109/IROS58592.2024.10802377},
	urldate = {2026-01-06},
	booktitle = {2024 {IEEE}/{RSJ} {International} {Conference} on {Intelligent} {Robots} and {Systems} ({IROS})},
	publisher = {IEEE},
	author = {Pei, Guanran and Stella, Francesco and Meebed, Omar and Bing, Zhenshan and Santina, Cosimo Della and Hughes, Josie},
	month = oct,
	year = {2024},
	pages = {1847--1852},
}

@article{an_shape_2024,
	title = {Shape reconstruction of soft continuum robots via the fusion of local strains and global poses},
	volume = {5},
	issn = {26663864},
	url = {https://linkinghub.elsevier.com/retrieve/pii/S2666386424005174},
	doi = {10.1016/j.xcrp.2024.102224},
	number = {10},
	urldate = {2026-01-06},
	journal = {Cell Reports Physical Science},
	author = {An, Xin and Cui, Yafeng and Dong, Xuguang and Wang, Yixin and Du, Boyuan and Liu, Xin-Jun and Zhao, Huichan},
	month = oct,
	year = {2024},
	pages = {102224},
}

@article{thuruthel_model-based_2019,
	title = {Model-{Based} {Reinforcement} {Learning} for {Closed}-{Loop} {Dynamic} {Control} of {Soft} {Robotic} {Manipulators}},
	volume = {35},
	copyright = {https://ieeexplore.ieee.org/Xplorehelp/downloads/license-information/IEEE.html},
	issn = {1552-3098, 1941-0468},
	url = {https://ieeexplore.ieee.org/document/8531756/},
	doi = {10.1109/TRO.2018.2878318},
	number = {1},
	urldate = {2026-01-06},
	journal = {IEEE Transactions on Robotics},
	author = {Thuruthel, Thomas George and Falotico, Egidio and Renda, Federico and Laschi, Cecilia},
	month = feb,
	year = {2019},
	pages = {124--134},
}

@article{rucker_statics_2011,
	title = {Statics and {Dynamics} of {Continuum} {Robots} {With} {General} {Tendon} {Routing} and {External} {Loading}},
	volume = {27},
	copyright = {https://ieeexplore.ieee.org/Xplorehelp/downloads/license-information/IEEE.html},
	issn = {1552-3098},
	url = {http://ieeexplore.ieee.org/document/5957337/},
	doi = {10.1109/TRO.2011.2160469},
	number = {6},
	urldate = {2026-01-06},
	journal = {IEEE Transactions on Robotics},
	author = {Rucker, D. Caleb and Webster III, Robert J.},
	month = dec,
	year = {2011},
	pages = {1033--1044},
}

@article{till_real-time_2019,
	title = {Real-time dynamics of soft and continuum robots based on {Cosserat} rod models},
	volume = {38},
	issn = {0278-3649, 1741-3176},
	url = {https://journals.sagepub.com/doi/10.1177/0278364919842269},
	doi = {10.1177/0278364919842269},
	number = {6},
	urldate = {2026-01-06},
	journal = {The International Journal of Robotics Research},
	author = {Till, John and Aloi, Vincent and Rucker, Caleb},
	month = may,
	year = {2019},
	pages = {723--746},
}

@inproceedings{duriez_control_2013,
	address = {Karlsruhe, Germany},
	title = {Control of elastic soft robots based on real-time finite element method},
	isbn = {978-1-4673-5643-5 978-1-4673-5641-1},
	url = {http://ieeexplore.ieee.org/document/6631138/},
	doi = {10.1109/ICRA.2013.6631138},
	urldate = {2026-01-06},
	booktitle = {2013 {IEEE} {International} {Conference} on {Robotics} and {Automation}},
	publisher = {IEEE},
	author = {Duriez, Christian},
	month = may,
	year = {2013},
	pages = {3982--3987},
}

@article{renda_dynamic_2014,
	title = {Dynamic {Model} of a {Multibending} {Soft} {Robot} {Arm} {Driven} by {Cables}},
	volume = {30},
	copyright = {https://ieeexplore.ieee.org/Xplorehelp/downloads/license-information/IEEE.html},
	issn = {1552-3098, 1941-0468},
	url = {http://ieeexplore.ieee.org/document/6827980/},
	doi = {10.1109/TRO.2014.2325992},
	number = {5},
	urldate = {2026-01-06},
	journal = {IEEE Transactions on Robotics},
	author = {Renda, Federico and Giorelli, Michele and Calisti, Marcello and Cianchetti, Matteo and Laschi, Cecilia},
	month = oct,
	year = {2014},
	pages = {1109--1122},
}

@article{goury_fast_2018,
	title = {Fast, {Generic}, and {Reliable} {Control} and {Simulation} of {Soft} {Robots} {Using} {Model} {Order} {Reduction}},
	volume = {34},
	copyright = {https://ieeexplore.ieee.org/Xplorehelp/downloads/license-information/IEEE.html},
	issn = {1552-3098, 1941-0468},
	url = {https://ieeexplore.ieee.org/document/8453914/},
	doi = {10.1109/TRO.2018.2861900},
	number = {6},
	urldate = {2026-01-06},
	journal = {IEEE Transactions on Robotics},
	author = {Goury, Olivier and Duriez, Christian},
	month = dec,
	year = {2018},
	pages = {1565--1576},
}

@article{bao_kinematics_2019,
	title = {Kinematics {Modeling} of a {Twisted} and {Coiled} {Polymer}-{Based} {Elastomer} {Soft} {Robot}},
	volume = {7},
	copyright = {https://creativecommons.org/licenses/by/4.0/legalcode},
	issn = {2169-3536},
	url = {https://ieeexplore.ieee.org/document/8844706/},
	doi = {10.1109/ACCESS.2019.2942486},
	urldate = {2026-01-06},
	journal = {IEEE Access},
	author = {Bao, Jiali and Chen, Weihang and Xu, Jing},
	year = {2019},
	pages = {136792--136800},
}

@article{bruder_data-driven_2021,
	title = {Data-{Driven} {Control} of {Soft} {Robots} {Using} {Koopman} {Operator} {Theory}},
	volume = {37},
	copyright = {https://ieeexplore.ieee.org/Xplorehelp/downloads/license-information/IEEE.html},
	issn = {1552-3098, 1941-0468},
	url = {https://ieeexplore.ieee.org/document/9277915/},
	doi = {10.1109/TRO.2020.3038693},
	number = {3},
	urldate = {2026-01-06},
	journal = {IEEE Transactions on Robotics},
	author = {Bruder, Daniel and Fu, Xun and Gillespie, R. Brent and Remy, C. David and Vasudevan, Ram},
	month = jun,
	year = {2021},
	pages = {948--961},
}

@inproceedings{singh_controlling_2023,
	address = {San Diego, CA, USA},
	title = {Controlling the {Shape} of {Soft} {Robots} {Using} the {Koopman} {Operator}},
	copyright = {https://doi.org/10.15223/policy-029},
	isbn = {979-8-3503-2806-6},
	url = {https://ieeexplore.ieee.org/document/10156145/},
	doi = {10.23919/ACC55779.2023.10156145},
	urldate = {2026-01-06},
	booktitle = {2023 {American} {Control} {Conference} ({ACC})},
	publisher = {IEEE},
	author = {Singh, Ajai and Sun, Jiefeng and Zhao, Jianguo},
	month = may,
	year = {2023},
	pages = {153--158},
}

@article{korda_linear_2018,
	title = {Linear predictors for nonlinear dynamical systems: {Koopman} operator meets model predictive control},
	volume = {93},
	issn = {00051098},
	shorttitle = {Linear predictors for nonlinear dynamical systems},
	url = {https://linkinghub.elsevier.com/retrieve/pii/S000510981830133X},
	doi = {10.1016/j.automatica.2018.03.046},
	urldate = {2026-01-06},
	journal = {Automatica},
	author = {Korda, Milan and Mezić, Igor},
	month = jul,
	year = {2018},
	pages = {149--160},
}

@article{haggerty_control_2023,
	title = {Control of soft robots with inertial dynamics},
	volume = {8},
	issn = {2470-9476},
	url = {https://www.science.org/doi/10.1126/scirobotics.add6864},
	doi = {10.1126/scirobotics.add6864},
	number = {81},
	urldate = {2026-01-06},
	journal = {Science Robotics},
	author = {Haggerty, David A. and Banks, Michael J. and Kamenar, Ervin and Cao, Alan B. and Curtis, Patrick C. and Mezić, Igor and Hawkes, Elliot W.},
	month = aug,
	year = {2023},
	pages = {eadd6864},
}

@inproceedings{bruder_modeling_2019,
	title = {Modeling and {Control} of {Soft} {Robots} {Using} the {Koopman} {Operator} and {Model} {Predictive} {Control}},
	isbn = {978-0-9923747-5-4},
	url = {http://www.roboticsproceedings.org/rss15/p60.pdf},
	doi = {10.15607/RSS.2019.XV.060},
	urldate = {2026-01-06},
	booktitle = {Robotics: {Science} and {Systems} {XV}},
	publisher = {Robotics: Science and Systems Foundation},
	author = {Bruder, Daniel and Gillespie, Brent and David Remy, C. and Vasudevan, Ram},
	month = jun,
	year = {2019},
}

@article{junfeng_shape_2023,
	title = {Shape {Control} of a {Dual}-{Segment} {Soft} {Robot} using {Depth} {Vision}},
	volume = {14},
	issn = {21565570, 2158107X},
	url = {http://thesai.org/Publications/ViewPaper?Volume=14&Issue=6&Code=IJACSA&SerialNo=4},
	doi = {10.14569/IJACSA.2023.0140604},
	number = {6},
	urldate = {2026-01-08},
	journal = {International Journal of Advanced Computer Science and Applications},
	author = {Junfeng, Hu and Jun, Zhang},
	year = {2023},
}

@misc{kasaei_shape-aware_2025,
	title = {Shape-{Aware} {Whole}-{Body} {Control} for {Continuum} {Robots} with {Application} in {Endoluminal} {Surgical} {Robotics}},
	url = {http://arxiv.org/abs/2510.12332},
	doi = {10.48550/arXiv.2510.12332},
	language = {en},
	urldate = {2026-01-08},
	publisher = {arXiv},
	author = {Kasaei, Mohammadreza and Ghobadi, Mostafa and Khadem, Mohsen},
	month = oct,
	year = {2025},
	note = {arXiv:2510.12332 [cs]},
}

@article{tang_general_2026,
	title = {A general soft robotic controller inspired by neuronal structural and plastic synapses that adapts to diverse arms, tasks, and perturbations},
	language = {en},
	journal = {Science AdvAnceS},
	author = {Tang, Zhiqiang and Tian, Liying and Xin, Wenci and Wang, Qianqian and Rus, Daniela and Laschi, Cecilia},
	year = {2026},
}

@article{fischer_dynamic_2023,
	title = {Dynamic {Task} {Space} {Control} {Enables} {Soft} {Manipulators} to {Perform} {Real}‐{World} {Tasks}},
	volume = {5},
	issn = {2640-4567, 2640-4567},
	url = {https://advanced.onlinelibrary.wiley.com/doi/10.1002/aisy.202200024},
	doi = {10.1002/aisy.202200024},
	number = {1},
	urldate = {2026-01-08},
	journal = {Advanced Intelligent Systems},
	author = {Fischer, Oliver and Toshimitsu, Yasunori and Kazemipour, Amirhossein and Katzschmann, Robert K.},
	month = jan,
	year = {2023},
	pages = {2200024},
}

@article{proctor_generalizing_2018,
	title = {Generalizing {Koopman} {Theory} to {Allow} for {Inputs} and {Control}},
	volume = {17},
	url = {https://epubs.siam.org/doi/10.1137/16M1062296},
	doi = {10.1137/16M1062296},
	number = {1},
	urldate = {2024-06-29},
	journal = {SIAM Journal on Applied Dynamical Systems},
	author = {Proctor, Joshua L. and Brunton, Steven L. and Kutz, J. Nathan},
	month = jan,
	year = {2018},
	pages = {909--930},
}

@article{kaiser_data-driven_2021,
	title = {Data-driven discovery of {Koopman} eigenfunctions for control},
	volume = {2},
	issn = {2632-2153},
	url = {https://iopscience.iop.org/article/10.1088/2632-2153/abf0f5},
	doi = {10.1088/2632-2153/abf0f5},
	number = {3},
	urldate = {2024-06-29},
	journal = {Machine Learning: Science and Technology},
	author = {Kaiser, Eurika and Kutz, J Nathan and Brunton, Steven L},
	month = sep,
	year = {2021},
	pages = {035023},
}

@article{williams_datadriven_2015,
	title = {A {Data}–{Driven} {Approximation} of the {Koopman} {Operator}: {Extending} {Dynamic} {Mode} {Decomposition}},
	volume = {25},
	issn = {1432-1467},
	shorttitle = {A {Data}–{Driven} {Approximation} of the {Koopman} {Operator}},
	url = {https://doi.org/10.1007/s00332-015-9258-5},
	doi = {10.1007/s00332-015-9258-5},
	language = {en},
	number = {6},
	urldate = {2024-06-30},
	journal = {Journal of Nonlinear Science},
	author = {Williams, Matthew O. and Kevrekidis, Ioannis G. and Rowley, Clarence W.},
	month = dec,
	year = {2015},
	pages = {1307--1346},
}

@article{seheult_robust_1989,
	title = {Robust {Regression} and {Outlier} {Detection}.},
	volume = {152},
	doi = {10.2307/2982847},
	journal = {Journal of the Royal Statistical Society. Series A (Statistics in Society)},
	author = {Seheult, Allan and Green, P. and Rousseeuw, Peter and Leroy, Annick},
	month = jan,
	year = {1989},
	pages = {133},
}

@article{tibshirani_regression_1996,
	title = {Regression {Shrinkage} and {Selection} via the {Lasso}},
	volume = {58},
	issn = {0035-9246},
	url = {https://www.jstor.org/stable/2346178},
	number = {1},
	urldate = {2024-07-10},
	journal = {Journal of the Royal Statistical Society. Series B (Methodological)},
	author = {Tibshirani, Robert},
	year = {1996},
	pages = {267--288},
}

@article{stellato_osqp_2020,
	title = {{OSQP}: an operator splitting solver for quadratic programs},
	volume = {12},
	issn = {1867-2957},
	shorttitle = {{OSQP}},
	url = {https://doi.org/10.1007/s12532-020-00179-2},
	doi = {10.1007/s12532-020-00179-2},
	language = {en},
	number = {4},
	urldate = {2025-01-12},
	journal = {Mathematical Programming Computation},
	author = {Stellato, Bartolomeo and Banjac, Goran and Goulart, Paul and Bemporad, Alberto and Boyd, Stephen},
	month = dec,
	year = {2020},
	pages = {637--672},
}

@inproceedings{dewi_lightweight_2024,
	address = {San Diego, CA, USA},
	title = {A {Lightweight} {Modular} {Segment} {Design} for {Tendon}-{Driven} {Continuum} {Robots} with {Pre}-{Programmable} {Stiffness}},
	copyright = {https://doi.org/10.15223/policy-029},
	isbn = {979-8-3503-8181-8},
	url = {https://ieeexplore.ieee.org/document/10522016/},
	doi = {10.1109/RoboSoft60065.2024.10522016},
	language = {en},
	urldate = {2026-01-14},
	booktitle = {2024 {IEEE} 7th {International} {Conference} on {Soft} {Robotics} ({RoboSoft})},
	publisher = {IEEE},
	author = {Dewi, Puspita Triana and Rao, Priyanka and Burgner-Kahrs, Jessica},
	month = apr,
	year = {2024},
	pages = {531--536},
}

@article{guan_trimmed_2023,
	title = {Trimmed helicoids: an architectured soft structure yielding soft robots with high precision, large workspace, and compliant interactions},
	volume = {1},
	issn = {2731-4278},
	shorttitle = {Trimmed helicoids},
	url = {https://www.nature.com/articles/s44182-023-00004-7},
	doi = {10.1038/s44182-023-00004-7},
	language = {en},
	number = {1},
	urldate = {2026-01-14},
	journal = {npj Robotics},
	author = {Guan, Qinghua and Stella, Francesco and Della Santina, Cosimo and Leng, Jinsong and Hughes, Josie},
	month = oct,
	year = {2023},
	pages = {4},
}

@article{della_santina_improved_2020,
  author  = {Della Santina, Cosimo and Rus, Daniela},
  title   = {On an {Improved} {State} {Parametrization} for {Soft} {Robots} with {Piecewise} {Constant} {Curvature} and Its Use in {Model}-Based Control},
  journal = {IEEE Robotics and Automation Letters},
  year    = {2020},
  volume  = {5},
  number  = {2},
  pages   = {1001--1008},
  month   = apr,
  doi     = {10.1109/LRA.2020.2969930}
}

@inproceedings{katzschmann_dynamic_2019,
	address = {Seoul, Korea (South)},
	title = {Dynamic {Motion} {Control} of {Multi}-{Segment} {Soft} {Robots} {Using} {Piecewise} {Constant} {Curvature} {Matched} with an {Augmented} {Rigid} {Body} {Model}},
	copyright = {https://ieeexplore.ieee.org/Xplorehelp/downloads/license-information/IEEE.html},
	isbn = {978-1-5386-9260-8},
	url = {https://ieeexplore.ieee.org/document/8722799/},
	doi = {10.1109/ROBOSOFT.2019.8722799},
	urldate = {2026-03-04},
	booktitle = {2019 2nd {IEEE} {International} {Conference} on {Soft} {Robotics} ({RoboSoft})},
	publisher = {IEEE},
	author = {Katzschmann, Robert K. and Santina, Cosimo Della and Toshimitsu, Yasunori and Bicchi, Antonio and Rus, Daniela},
	month = apr,
	year = {2019},
	pages = {454--461},
}

@article{shi_koopman_2025,
	title = {Koopman {Operators} in {Robot} {Learning}},
    doi={10.1109/TRO.2026.3654384},
	language = {en},
	urldate = {2026-03-04},
    author={Shi, Lu and Haseli, Masih and Mamakoukas, Giorgos and Bruder, Daniel and Abraham, Ian and Murphey, Todd and Cortés, Jorge and Karydis, Konstantinos},
    journal={IEEE Transactions on Robotics}, 
	volume={42},
	year={2026},
	pages={1088-1107},
}

@article{veil_shape-space_2026,
  author  = {Veil, Carina and Flaschel, Moritz and Kuhl, Ellen},
  title   = {Shape-Space Graphs: Fast and Collision-Free Path Planning for Soft Robots},
  journal = {IEEE Robotics and Automation Letters},
  year    = {2026},
  volume  = {11},
  number  = {5},
  pages   = {5582--5589},
  month   = may,
  doi     = {10.1109/LRA.2026.11433777}
}

@article{liu_path_2023,
	title = {Path planning with obstacle avoidance for soft robots based on improved particle swarm optimization algorithm},
	volume = {3},
	issn = {2770-3541, 2770-3541},
	url = {https://www.oaepublish.com/articles/ir.2023.31},
	doi = {10.20517/ir.2023.31},
	number = {4},
	urldate = {2026-03-12},
	journal = {Intelligence \& Robotics},
	author = {Liu, Hongwei and Jiang, Yang and Liu, Manlu and Zhang, Xinbin and Huo, Jianwen and Su, Haoxiang},
	month = oct,
	year = {2023},
	pages = {565--80},
}

@ARTICLE{wang_dataefficient_2026,
  author={Wang, Jiahe and Ristich, Eron and Weissman, Eric and Ren, Yi and Sun, Jiefeng},
  journal={IEEE Robotics and Automation Letters}, 
  title={Data-Efficient Real-Time Control of an Artificial-Muscle-Driven Continuum Robot With Physics-Informed Koopman Operator}, 
  year={2026},
  volume={11},
  number={4},
  pages={5080-5087},
  }

@article{george_thuruthel_control_2018,
	title = {Control {Strategies} for {Soft} {Robotic} {Manipulators}: {A} {Survey}},
	volume = {5},
	issn = {2169-5172, 2169-5180},
	shorttitle = {Control {Strategies} for {Soft} {Robotic} {Manipulators}},
	url = {https://journals.sagepub.com/doi/10.1089/soro.2017.0007},
	number = {2},
	urldate = {2026-03-25},
	journal = {Soft Robotics},
	author = {George Thuruthel, Thomas and Ansari, Yasmin and Falotico, Egidio and Laschi, Cecilia},
	month = apr,
	year = {2018},
	pages = {149--163},
}

@article{mbakop_parametric_2022,
	title = {Parametric {PH} {Curves} {Model} {Based} {Kinematic} {Control} of the {Shape} of {Mobile} {Soft} {Manipulators} in {Unstructured} {Environment}},
	volume = {69},
	issn = {0278-0046, 1557-9948},
	language = {en},
	number = {10},
	urldate = {2026-03-27},
	journal = {IEEE Transactions on Industrial Electronics},
	author = {Mbakop, Steeve and Tagne, Gilles and Drakunov, Sergey and Merzouki, Rochdi},
	month = oct,
	year = {2022},
	pages = {10292--10300},
}

@article{yang_combined_2024,
	title = {A combined kinodynamic motion planning method for multisegment continuum manipulators in confined spaces},
	volume = {112},
	issn = {0924-090X, 1573-269X},
	url = {https://link.springer.com/10.1007/s11071-023-09190-3},
	doi = {10.1007/s11071-023-09190-3},
	pages = {2721--2744},
	number = {4},
	journal = {Nonlinear Dynamics},
	shortjournal = {Nonlinear Dyn},
	author = {Yang, Jinzhao and Peng, Haijun and Wu, Shunan and Zhang, Jie and Wu, Zhigang and Wu, Jianing},
	date = {2024-02},
}

@article{yuan_hybrid_2026,
	title = {Hybrid Offline–Online Configuration Planning Approach for Continuum Robots Based on Real-Time Shape Estimation},
	volume = {26},
	issn = {1424-8220},
	url = {https://www.mdpi.com/1424-8220/26/4/1129},
	doi = {10.3390/s26041129},
	pages = {1129},
	number = {4},
	journal = {Sensors},
	shortjournal = {Sensors},
	author = {Yuan, Hexiang and Jing, Zhibo and He, Yibo and Han, Jianda and Zhang, Juanjuan},
	date = {2026-02-10},
	langid = {english},
}

@article{wu_novel_2022,
	title = {A novel obstacle avoidance heuristic algorithm of continuum robot based on {FABRIK}},
	volume = {65},
	issn = {1674-7321, 1869-1900},
	url = {https://link.springer.com/10.1007/s11431-022-2179-9},
	doi = {10.1007/s11431-022-2179-9},
	pages = {2952--2966},
	number = {12},
	journal = {Science China Technological Sciences},
	shortjournal = {Sci. China Technol. Sci.},
	author = {Wu, {HaoRan} and Yu, {JingJun} and Pan, Jie and Pei, Xu},
	date = {2022-12},
}

@article{shentu_sampling-based_2026,
	title = {Sampling-Based Follow-the-Leader Motion Planning for Manipulator-Mounted Continuum Robots},
	url = {http://arxiv.org/abs/2605.11618},
	doi = {10.48550/arXiv.2605.11618},
	number = {{arXiv}:2605.11618},
	publisher = {{arXiv}},
	author = {Shentu, Chengnan and Baldassini, Nicholas and Iseoluwa, Oluwagbotemi D. and Gondokaryono, Radian and Burgner-Kahrs, Jessica},
	date = {2026-05-12},
	eprinttype = {arxiv},
	eprint = {2605.11618 [cs.RO]},
}

\end{document}